%% file: main.tex
\documentclass[10pt,logo,copyright]{nvidiatechreport}
\input{packages}

\input{common}

\usepackage{pifont}
\usepackage{diagbox}

\usepackage[nameinlink]{cleveref}
\crefname{equation}{Eq.}{Eqs.}
\crefname{figure}{Fig.}{Figs.}
\crefname{section}{Sec.}{Sec.}
\crefname{appendix}{App.}{App.}
\crefname{table}{Tab.}{Tabs.}
\crefname{algorithm}{Algo}{Algo}
\crefname{thm}{Thm}{Thm}
\Crefname{thm}{Thm}{Thm}
\crefname{prop}{Prop}{Prop}

\definecolor{darkred}{rgb}{0.7, 0.0, 0.0}

\newcommand{\crefnames}[3]{%
  \@for\next:=#1\do{%
    \expandafter\crefname\expandafter{\next}{#2}{#3}%
  }%
}

\title{\textit{Causal-rCM}: A Unified Teacher-Forcing and Self-Forcing Open Recipe for Autoregressive Diffusion Distillation in Streaming Video Generation and Interactive World Models}
\shorttitle{\textit{Causal-rCM}: A Unified Teacher-Forcing and Self-Forcing Open Recipe for Autoregressive Diffusion Distillation}
\usepackage[symbol,stable]{footmisc}

\author{%
  Kaiwen Zheng$^{1,3}$, Guande He$^{2}$, Min Zhao$^{1}$, Jintao Zhang$^{1}$, Huayu Chen$^{1}$,\\
  \textbf{\Authfont Jianfei Chen$^{1}$, Chen-Hsuan Lin$^{3}$, Ming-Yu Liu$^{3}$, Jun Zhu$^{1*}$, Qianli Ma$^{3}$}\\~\\
  $^1$Tsinghua University \quad $^2$UT Austin \quad
  $^3$NVIDIA \\
  $^*$Corresponding Author \\
  \texttt{zkwthu@gmail.com;}\quad
\texttt{dcszj@tsinghua.edu.cn;}\quad\texttt{mingyul@nvidia.com} \\
  \url{https://github.com/NVlabs/rcm}
}

\begin{abstract}
\end{abstract}

\begin{document}
\maketitle
\abscontent
Autoregressive video diffusion with causal diffusion transformers has emerged as a major paradigm for real-time streaming video generation and action-conditioned interactive world models. In this work, we extend rCM, an advanced diffusion distillation framework, to autoregressive video diffusion. The core philosophy of rCM lies in the complementarity between forward and reverse divergences, represented by consistency models (CMs) and distribution matching distillation (DMD), respectively, in diffusion distillation. This philosophy naturally carries over to the autoregressive setting, where teacher-forcing (TF) provides an offline, forward-divergence causal training paradigm, while self-forcing (SF) corresponds to an on-policy, reverse-divergence refinement.

Our contributions are: (1) through extensive experiments, we show that teacher-forcing CM is currently the best complement to self-forcing DMD as an initialization strategy (2) we present the first implementation of teacher-forcing-based continuous-time CMs (e.g., sCM/MeanFlow) for autoregressive video diffusion, enabled by our custom-mask FlashAttention-2 JVP kernel, achieving 10$\times$ faster convergence compared to discrete-time CMs (dCMs) (3) we introduce Causal-rCM, a leading, unified, and scalable algorithm-infrastructure open recipe for diffusion distillation and causal training (4) we achieve state-of-the-art streaming video generation performance in both frame-wise and chunk-wise settings, using only synthetic data for training.

Notably, our distilled 2-step causal Wan2.1-1.3B model achieves a VBench-T2V score of 84.63 with only 1 or 2 sampling steps. We further apply Causal-rCM to Cosmos 3, an advanced omnimodal world foundation model for physical AI with action-conditioned generation capability, enabling an interactive world model.

\input{figures/vbench}

\tableofcontents

\section{Introduction}

Video diffusion models are widely recognized as a form of world simulators~\citep{brooks2024video,bao2024vidu,kong2024hunyuanvideo,wan2025,ali2025world,gao2025seedance,seedance2026seedance,nvidia2026cosmos3}. Instead of denoising all frames jointly with a bidirectional-attention diffusion transformer, autoregressive (AR) video diffusion~\citep{jin2025pyramidal,teng2025magi,chen2025skyreels} performs \textit{next-frame} or \textit{next-chunk} prediction with causal-attention diffusion transformers. This mirrors the shift from masked diffusion~\citep{sahoo2024simple,shi2024simplified,zheng2025masked} to block diffusion~\citep{arriola2025block} in the discrete diffusion regime. In this paradigm, the model is autoregressive across frames or chunks, while diffusion denoising is performed within each frame or chunk. This enables streaming long video generation~\citep{huang2025live,longlive,longlive_2.0}, interactive world models~\citep{hong2025relic,hunyuanworld2025hy,he2025matrixgame,gao2026advancing}, and embodied AR video diffusion for closed-loop robot control~\citep{feng2025vidarc,li2026causal,ye2026world}.

Common causal training paradigms, such as teacher-forcing (TF) and diffusion-forcing (DF)~\citep{chen2024diffusion}, suffer from error accumulation and quality degradation over time during AR diffusion inference, commonly known as exposure bias~\citep{schmidt2019generalization,ning2024elucidating}. The recent self-forcing paradigm~\citep{huang2025selfforcing,lin2025autoregressive} resolves this issue by using on-policy training to tackle the training-inference gap, coupled with distribution matching distillation (DMD)~\citep{yin2024one,yin2024improved} or adversarial GAN losses~\citep{lin2025diffusion} for diffusion step distillation. Self-forcing approaches have pushed AR video diffusion toward practical low-latency, real-time, and long-horizon generation in streaming and interactive settings.

\input{figures/overview}

However, self-forcing with DMD or GAN objectives is sensitive to initialization and suffers from mode collapse, as DMD-style objectives are based on reverse-KL divergence and optimize student-generated rollouts. Existing AR diffusion systems therefore introduce different initialization strategies before self-forcing, such as ODE-pair regression~\citep{yin2025slow,huang2025selfforcing,he2025matrixgame,zhu2026causal}, diffusion-forcing-style causal adaptation~\citep{huang2025live,gao2026advancing}, or hybrid TF/DF initialization~\citep{hong2025relic}. These designs suggest that a stable offline causal objective is crucial before on-policy distribution matching, but the connection between initialization, causal training paradigms, and distillation losses remains underexplored.

In this work, we introduce \textbf{Causal-rCM}, extending rCM (score-regularized consistency model)~\citep{zheng2025large} to AR video diffusion. In rCM, the key insight is the forward-reverse complementarity at the level of distillation objectives: CMs act as forward-divergence, trajectory-preserving objectives, while DMD acts as a reverse-divergence, distribution-matching objective. In AR diffusion, an analogous complementarity arises at the level of causal training paradigms, where teacher-forcing provides an offline, mode-covering training signal and self-forcing provides an on-policy refinement signal under autoregressive rollouts. Based on this correspondence, Causal-rCM uses teacher-forcing CM for few-step causal distillation on offline causal contexts and teacher trajectories, and self-forcing DMD to directly optimize the inference-time few-step distribution.

\paragraph{Relation to Prior Art} CMs are widely used as initialization or regularization for DMD- and GAN-based diffusion distillation~\citep{lin2025diffusion,zheng2025large}. Notably, for AR diffusion, APT2~\citep{lin2025autoregressive} has adopted teacher-forcing-based CM as initialization for the self-forcing stage, with later theoretical support through the lens of frame-level injectivity~\citep{zhu2026causal}. Causal-rCM differs from previous works by (1) providing a unified divergence perspective on different causal training paradigms, distillation losses, and their synergy, echoing the high-level principle of rCM; (2) conducting a holistic and systematic investigation of different initialization strategies for self-forcing DMD, uncovering their pros and cons; (3) providing the first implementation of teacher-forcing based continuous-time consistency models (sCM~\citep{lu2024simplifying}, MeanFlow~\citep{geng2025mean}) with our custom-mask FlashAttention-2 JVP kernel, achieving 10$\times$ faster convergence compared to discrete-time CMs (dCMs); (4) introducing a leading, unified, and scalable algorithm-infrastructure open recipe for diffusion distillation and causal training, achieving state-of-the-art performance in AR diffusion distillation.

\paragraph{Forward-Reverse Objective Complementarity}

\input{tables/comparison}

The broader philosophy of jointly leveraging forward and reverse objectives has appeared across diffusion mid-training, diffusion reinforcement learning, and diffusion distillation. Forward or offline objectives, such as diffusion losses, teacher-forcing losses, and CM losses on real data or teacher trajectories, provide stable training signals and preserve mode coverage. Reverse or on-policy objectives, such as DMD, adversarial losses, and reward-driven optimization on generated samples, directly improve the generated distribution but are more sensitive to initialization and coverage. As summarized in Table~\ref{tab:forward_reverse_complementarity}, recent methods including DDO~\citep{zheng2025direct}, DiffusionNFT~\citep{zheng2025diffusionnft}, DDRL~\citep{ye2025data}, and rCM~\citep{zheng2025large} all benefit from this complementarity. Causal-rCM instantiates the same principle in AR diffusion distillation: teacher-forcing CM serves as the forward/offline component, while self-forcing DMD serves as the reverse/on-policy component.

\section{Background}
\subsection{Diffusion Models}
\textbf{Diffusion models (DMs)}~\citep{ho2020denoising,song2020score} learn continuous data distributions by gradually perturbing clean data $\x_0\sim p_\text{data}$ with Gaussian noise, which generates a trajectory $\{\x_t\}_{t=0}^T$ along with associated marginals $\{q_t\}_{t=0}^T$, and then learning to reverse this process. The forward process follows a closed-form transition kernel $q_{t|0}(\x_t|\x_0)=\Nc(\alpha_t\x_0,\sigma_t^2\Iv)$ with predefined noise schedule $\alpha_t,\sigma_t$, enabling reparameterization as $\x_t=\alpha_t\x_0+\sigma_t\epsilonv,\epsilonv\sim\Nc(\vect 0,\Iv)$. The sampling process of DMs can follow the probability flow ordinary differential equation (PF-ODE) $\dm\x_t = \left[f(t)\x_t - \frac{1}{2}g^2(t)\nabla_{\x_t} \log q_t(\x_t)\right] \dm t$, where $f(t)=\frac{\mathrm{d}\log \alpha_t}{\mathrm{d} t}$, $g^2(t)=\frac{\mathrm{d} \sigma_t^2}{\mathrm{d} t}-2\frac{\mathrm{d}\log \alpha_t}{\mathrm{d} t}\sigma_t^2$, and $\nabla_{\x_t} \log q_t(\x_t)$ is the \textit{score function}~\citep{song2020score}. A key property of DMs is the theoretical equivalence of different parameterizations: the network may predict the score ($\nabla_{\x_t}\log q_t(\x_t)$), the noise ($\epsilonv$), the clean data ($\x_0$), or the velocity ($\vv$), with optimal predictors being analytically interconvertible~\citep{zheng2023improved}. With velocity parameterization $\vv_\theta$~\citep{zheng2023improved}, DMs are trained by minimizing the mean square error (MSE) $\E_{\x_0\sim p_\text{data},\epsilonv,t}[w(t)\|\vv_\theta(\x_t,t)-\vv\|_2^2]$, where the regression target is $\vv=\dot{\alpha}_t \x_0 + \dot{\sigma}_t \epsilonv$ (denote $\dot{f}_t\coloneq\mathrm d f_t/\mathrm dt$), and the PF-ODE is simplified to $\frac{\dm\x_t}{\dm t}=\vv_\theta(\x_t,t)$, commonly known as flow matching~\citep{lipman2022flow}. A notable special case, rectified flow (RF)~\citep{liu2022flow}, employs the schedule $\alpha_t=1-t,\sigma_t=t$, which simplifies the velocity target to $\vv=\epsilonv-\x_0$. 
\subsection{Diffusion Distillation}
\paragraph{Consistency Distillation}
\textbf{Consistency models (CMs)} ~\citep{song2023consistency} aim to learn a \textit{consistency function} $\fv_\theta: (\xv_t, t) \mapsto \x_0$ which maps the point $\x_t$ at arbitrary time $t$ on the teacher PF-ODE trajectory to the initial point $\x_0$. Given a free-form student network $\Fv_\theta(\x,t)$, the consistency function is usually parameterized as
$
\fv_\theta(\x,t)=c_\text{skip}(t)\x+c_\text{out}(t)\Fv_\theta(c_\text{in}(t)\x,c_\text{noise}(t)),
$
with $c_\text{skip}(0)=1$ and $c_\text{out}(0)=0$ (e.g., $\fv_\theta(\x,t)=\x-t\Fv_\theta(\x,t)$ under the RF schedule). This parameterization naturally satisfies the boundary condition $\fv_{\theta}(\xv,0)\equiv \xv$. Here, $\fv_\theta$ is the direct counterpart of the data predictor (denoiser) in DMs, while $\Fv_\theta(\x,t)$ corresponds to the velocity predictor $\vv_\theta$.

The CM objective enforces consistent student outputs at adjacent timesteps $t-\Delta t$ and $t$ along the teacher trajectory. \textbf{Discrete-time CMs (dCMs)} minimize the following objective with $\Delta t>0$:
\begin{equation}
\label{eq:loss-cm-discrete}
\Lc_{\text{dCM}}(\theta)=\E_{\x_0\sim p_\text{data},\epsilonv,t}\left[w(t)d\left(\fv_{\theta}(\xv_t, t),\fv_{\theta^-}(\hat\x_{t-\Delta t}, t-\Delta t)\right)\right],
\end{equation}
where $w(\cdot)$ is a positive weighting function, $d(\cdot,\cdot)$ is a distance metric, $\theta^-$ is the stop-gradient version of $\theta$, and $\hat\x_{t-\Delta t}$ is obtained by solving the teacher PF-ODE from $(\x_t,t)$ to $t-\Delta t$ with numerical solvers.

\textbf{Continuous-time CMs (sCM)}~\citep{lu2024simplifying} take the limit $\Delta t\rightarrow 0$ in dCM to obtain a more accurate objective. When $d(\x,\y)=\|\x-\y\|_2^2$, the instantaneous‌ CM loss becomes
$\E_{\x_0\sim p_\text{data},\epsilonv,t}\left[w(t)\fv_{\theta}(\x_t, t)^\top\frac{\dm\fv_{\theta^-}(\x_t,t)}{\dm t}\right]$, where $\frac{\dm\fv_{\theta^-}(\x_t,t)}{\dm t}=\nabla_{\x_t}\fv_{\theta^-}(\x_t,t)\frac{\dm\x_t}{\dm t}+\partial_t\fv_{\theta^-}(\x_t,t)$ is the \textit{tangent} of $\fv_\theta$ at $(\x_t,t)$ along the teacher ODE trajectory $\frac{\dm\x_t}{\dm t}=\vv_{\text{teacher}}(\x_t,t)$. This tangent can be efficiently computed by forward-mode automatic differentiation, \textit{Jacobian-vector product (JVP)}: $\frac{\dm\fv_{\theta^-}(\x_t,t)}{\dm t}=\texttt{JVP}(\fv_{\theta^-}, (\x_t,t), (\frac{\dm\x_t}{\dm t}, 1))$. sCM further applies MSE reformulation and tangent normalization, reducing the loss to
\begin{equation}
    \Lc_{\text{sCM}}(\theta)=\E_{\x_0\sim p_\text{data},\epsilonv,t\sim p_G}\left[\left\|\Fv_\theta(\x_t,t)-\Fv_{\theta^-}(\x_t,t)-\frac{\gv}{\|\gv\|_2^2+c}\right\|_2^2\right],\quad \gv=w(t)\frac{\dm\fv_{\theta^-}(\x_t,t)}{\dm t}
\end{equation}

\textbf{MeanFlow}~\citep{geng2025mean} can be viewed as combining sCM with consistency trajectory models (CTMs)~\citep{kim2023consistency} under the RF schedule. CTMs extend CMs by adding another time condition $s<t$ and defining a \textit{consistency trajectory function} $\fv_\theta:(\xv_t,t,s)\mapsto \x_s$, which maps the point $\x_t$ to a less noisy point $\x_s$ on the teacher ODE trajectory. The infinitesimal jump from $t$ to $t$, i.e., $\fv_\theta(\x_t,t,t)$, reduces to the diffusion denoiser and serves as an anchor for applying the diffusion loss. This anchor enhances training stability, preserves multi-step sampling, and enables training few-step models from scratch. Thus, CTMs can be viewed as an interpolation between DMs and CMs. In the continuous-time case, CTMs can be optimized with an objective similar to sCM:
\begin{equation}
\begin{aligned}
    \Lc_{\text{sCTM}}(\theta)&=\E_{\x_0\sim p_\text{data},\epsilonv,t,s}\left[\left\|\Fv_\theta(\x_t,t,s)-\Fv_{\theta^-}(\x_t,t,s)-\frac{\gv}{\|\gv\|_2^2+c}\right\|_2^2\right],\quad \gv=w(t)\frac{\dm\fv_{\theta^-}(\x_t,t,s)}{\dm t}\\
    &=\E_{\x_0\sim p_\text{data},\epsilonv,t,s}\left[\frac{\|\Delta_\theta\|_2^2}{\|\Delta_{\theta^-}\|_2^2+c}\right],\quad \Delta_\theta=\Fv_\theta(\x_t,t,s)-\Fv_{\theta^-}(\x_t,t,s)-\frac{\dm\fv_{\theta^-}(\x_t,t,s)}{\dm t}
\end{aligned}
\end{equation}
Under the RF schedule, we have
$
    \fv_{\theta^-}(\x_t,t,s)\coloneq \x_t-(t-s)\Fv_{\theta^-}(\x_t,t,s)$ , and $
    \frac{\dm \fv_{\theta^-}(\x_t,t,s)}{\dm t}=\frac{\dm \x_t}{\dm t}-\Fv_{\theta^-}(\x_t,t,s)-(t-s)\frac{\dm \Fv_{\theta^-}(\x_t,t,s)}{\dm t}
$. Since $\Fv_\theta$ is the velocity predictor $\vv_\theta$, if we take
$\frac{\dm\x_t}{\dm t}$ as the ground-truth velocity $\vv=\epsilonv-\x_0$, then
\begin{equation}
    \Delta_\theta=\vv_\theta(\x_t,t,s)-\vv+(t-s)\texttt{JVP}(\vv_{\theta^-}, (\x_t,t,s), (\vv,1,0))
\end{equation}
which recovers the MeanFlow objective. Alternatively, we can set
$\frac{\dm\x_t}{\dm t}=\vv_{\text{teacher}}(\x_t,t)$ and use the same formulation for distillation, rather than training from scratch.
\paragraph{Distribution Matching Distillation}
\textbf{Distribution matching distillation (DMD)}~\citep{yin2024one,yin2024improved} is a simple and effective type of score distillation~\citep{wang2023prolificdreamer,zhou2024score}. 
Given a few-step student generator $\x_0^\theta=\Gv_\theta(\zv)$, $\zv\sim p(\zv)$ with prior distribution $p(\zv)$, DMD aims to match the student distribution $p_\theta$ with the teacher distribution $p_{\text{teacher}}$ by minimizing the \textit{reverse-KL divergence} on their diffused marginals:
\begin{equation}
\Lc_{\mathrm{DMD}\text{-raw}}(\theta)= \E_t\left[\kl{p_\theta^t}{p_{\text{teacher}}^t}\right],
\quad
\x_t \sim q_{t|0}(\x_t|\x_0^\theta).
\end{equation}
The gradient of this objective can be written as a score difference between the student and teacher distributions:
\begin{equation}
\nabla_\theta \Lc_{\mathrm{DMD}\text{-raw}}(\theta)
=
\E_{\zv,\epsilonv,t}
\left[
w(t)
\left(
\nabla_{\x_t}\log p_\theta^t(\x_t)
-
\nabla_{\x_t}\log p_{\text{teacher}}^t(\x_t)
\right)^\top
\frac{\dm \x_t}{\dm \theta}
\right].
\end{equation}
The teacher score is provided by the pretrained DM, while the student score is intractable for a few-step generator. DMD therefore trains an auxiliary \textit{fake score} network $\phi$ on student-generated samples with $
\Lc_{\mathrm{fake}}(\phi)
=
\E_{\zv,\epsilonv,t}
\left[
\lambda(t)
\left\|
\fv_{\phi}(\x_t,t)-\x_0^\theta
\right\|_2^2
\right]
$, which serves as a proxy for the student score. In denoiser parameterization, the score difference can be written, up to a time-dependent scalar absorbed into $w(t)$, as the difference between the fake and teacher denoisers. With the adaptive normalization trick in DMD, the student can be updated with the following stop-gradient MSE objective:
\begin{equation}
\label{eq:dmd}
    \Lc_{\mathrm{DMD}}(\theta)=\E_{\x_0^\theta\sim p_\theta,\epsilonv,t\sim p_D}\left[\left\|\x_0^\theta-\texttt{sg}\left[\x_0^\theta-\frac{\fv_{\mathrm{fake}}(\x_t,t)-\fv_{\mathrm{teacher}}(\x_t,t)}{\texttt{mean}(\texttt{abs}(\x_0^\theta-\fv_{\mathrm{teacher}}(\x_t,t)))}\right]\right\|_2^2\right]
\end{equation}
DMD alternates between student ($\Lc_{\mathrm{DMD}}(\theta)$) and critic ($\Lc_{\mathrm{fake}}(\phi)$) phases, forming an adversarial training dynamic similar to GANs.
\subsection{Autoregressive Video Diffusion}

\input{figures/background}

\textbf{Autoregressive (AR) video diffusion} factorizes video generation along the temporal dimension. Given a video latent sequence $\xv_0=[\xv_0^1,\ldots,\xv_0^N]$ divided into frames or chunks, an AR model generates each block conditioned on previous blocks:
$
p_\theta(\xv_0)=\prod_{i=1}^{N}p_\theta(\xv_0^i|\xv_0^{<i}).
$ Within each temporal block, the model $p_\theta(\xv_0^i|\xv_0^{<i})$ still performs diffusion denoising, e.g., under the RF schedule, $\xv_t^i=(1-t)\xv_0^i+t\epsilonv^i$ with velocity target $\vv^i=\epsilonv^i-\xv_0^i$.
Different from bidirectional video diffusion, which denoises all frames jointly with full temporal attention, AR video diffusion uses causal attention so that each frame or chunk only attends to past context. This enables KV caching like LLMs and makes the model naturally suitable for streaming and interactive generation.

Fig.~\ref{fig:background} illustrates the three causal training paradigms: \textbf{teacher-forcing (TF)}, \textbf{diffusion-forcing (DF)}, and \textbf{self-forcing (SF)}.

In TF, the model predicts the current noisy block while attending to clean ground-truth history, i.e., $\vv_\theta(\xv_t^i,t|\xv_0^{<i})$. TF is stable and parallelizable via a specific attention mask, but it creates a training-inference gap: during inference, the model must condition on its own generated history rather than ground-truth context. 

DF assigns independent noise levels to different frames or chunks and trains the model under a block-causal attention mask, i.e., $\vv_\theta(\xv_{t_i}^i,t_i|\xv_{t_{<i}}^{<i})$. This exposes the model to noisy histories and improves robustness. However, the training-inference gap remains: perturbing ground-truth videos with synthetic noise does not match the errors and artifacts accumulated from model-generated rollouts at inference.

SF directly simulates AR inference during training. The student rolls out chunks sequentially with KV caching, $\tilde{\xv}_0^i=\Gv_\theta(\zv^i|\tilde{\xv}_0^{<i})$, and the loss is applied to the self-generated video. Therefore, SF trains the model under its own inference-time context distribution, directly addressing the exposure bias induced by the training-inference gaps in TF and DF. SF must be combined with reverse-type on-policy objectives, such as DMD or GAN losses.

\section{Causal-rCM: A Leading, Unified and Scalable Algorithm-Infrastructure Open Recipe for Diffusion Distillation and Causal Training}

\subsection{Algorithms}
\input{figures/comparison}
To extend rCM to autoregressive diffusion, we pair its two distillation objectives (CM, DMD) with two causal training paradigms, teacher-forcing (TF) and self-forcing (SF), respectively. This preserves the forward-reverse correspondence of rCM in the autoregressive setting: TF-CM provides an offline, forward-type consistency objective, whereas SF-DMD provides an on-policy, reverse-type distribution-matching objective.

TF-CM requires an autoregressive diffusion teacher that is evaluated under the same clean-context setting as the student during TF-based distillation. Such a causal teacher can be trained from scratch, or adapted from a pretrained bidirectional diffusion model, with TF or DF. It is arguably more reasonable to use TF because it exposes the teacher to clean historical frames, matching the context distribution used in TF-based distillation. The CM component can be instantiated either as the simple dCM or as more advanced continuous-time variants such as sCM and MeanFlow. For SF-DMD, following prior work~\citep{huang2025selfforcing,lin2025autoregressive}, we use a bidirectional teacher and a bidirectional fake-score network to provide real and fake score estimates on self-generated rollouts (Fig.~\ref{fig:background}(c)) and apply the DMD loss (Eqn.~\ref{eq:dmd}).

Unlike rCM, which combines CM and DMD in a joint-training style, Causal-rCM applies TF-CM and SF-DMD sequentially. The full pipeline consists of three stages: (1) \textit{TF} converts the bidirectional diffusion model into an autoregressive diffusion model, which serves as both the causal teacher and the student initialization for the subsequent TF-CM stage; (2) \textit{TF-CM} distills the causal teacher into a few-step causal student, which serves as the student initialization for the subsequent SF-DMD stage; and (3) \textit{SF-DMD refinement} further optimizes the student on its own autoregressive rollouts, reducing the training-inference gap and exposure bias. As summarized in Fig.~\ref{fig:comparison}, Causal-rCM provides a simple and strong recipe that avoids cumbersome ODE-pair knowledge distillation (KD)~\citep{luhman2021knowledge} and GAN-style post-training, while introducing a novel TF-sCM implementation and achieving state-of-the-art performance.

\subsubsection{Teacher-Forcing, Teacher-Forcing dCM and Self-Forcing DMD}

The core operation of TF-based training is to replace a standard single-state forward with a packed causal forward over concatenated clean context and noisy targets. Concretely, for a velocity predictor, instead of evaluating $\vv_\theta(\xv_t,t)$, we evaluate
\begin{equation}
\left[
\vv_\theta
\left(
[\xv_0^{\mathrm{clean}},\xv_t^{\mathrm{noisy}}],
[\vect{0}^{\mathrm{clean}},\vect{t}^{\mathrm{noisy}}];
\Mv_{\mathrm{TF}}
\right)
\right]_{\mathrm{noisy}},
\end{equation}
where $\Mv_{\mathrm{TF}}$ is the TF attention mask, the clean part provides ground-truth causal context at timestep $0$, and the loss is applied only to the noisy part. The mask ensures that each noisy block attends only to its allowed clean history and its own noisy tokens, matching the TF pattern in Fig.~\ref{fig:background}. Such TF-mask attention can be implemented with custom-mask attention operators such as FlexAttention~\citep{dong2024flex} or MagiAttention~\citep{magiattention2025}. An alternative is a \textit{two-pass} implementation: first cache the clean tokens under a block-causal attention mask, and then perform a second forward pass in which noisy tokens attend to the cached clean context. However, this design requires the clean-token KV cache to be retained in the computational graph, making it less compatible with activation checkpointing and more memory-intensive.

With a diffusion regression target $\vv=\epsilonv-\xv_0$ under the RF schedule, the ordinary TF objective is
\begin{equation}
\Lc_{\mathrm{TF}}(\theta)
=
\E_{\xv_0,\epsilonv,t}
\left[
w(t)
\left\|
\left[
\vv_\theta
\left(
[\xv_0^{\mathrm{clean}},\xv_t^{\mathrm{noisy}}],
[\vect{0}^{\mathrm{clean}},\vect{t}^{\mathrm{noisy}}];
\Mv_{\mathrm{TF}}
\right)
\right]_{\mathrm{noisy}}
-
\vv
\right\|_2^2
\right].
\end{equation}
This gives a full-step causal diffusion model. For TF-dCM, the clean context remains fixed, while the noisy part is moved along the causal teacher PF-ODE trajectory. Let $\hat{\xv}_{t-\Delta t}^{\mathrm{noisy}}$ be obtained by solving the causal teacher ODE from $\xv_t^{\mathrm{noisy}}$ at $t$ to $t-\Delta t$ under the same TF mask. The student minimizes
\begin{equation}
\begin{aligned}
\Lc_{\mathrm{TF\text{-}dCM}}(\theta)
=
\E_{\xv_0,\epsilonv,t}
\Bigg[
w(t)
d\Bigg(
&\left[
\fv_\theta
\left(
[\xv_0^{\mathrm{clean}},\xv_t^{\mathrm{noisy}}],
[\vect{0}^{\mathrm{clean}},\vect{t}^{\mathrm{noisy}}];
\Mv_{\mathrm{TF}}
\right)
\right]_{\mathrm{noisy}},\\
\
&\left[
\fv_{\theta^-}
\left(
[\xv_0^{\mathrm{clean}},\hat{\xv}_{t-\Delta t}^{\mathrm{noisy}}],
[\vect{0}^{\mathrm{clean}},\vect{t-\Delta t}^{\mathrm{noisy}}];
\Mv_{\mathrm{TF}}
\right)
\right]_{\mathrm{noisy}}
\Bigg)
\Bigg].
\end{aligned}
\end{equation}

SF-DMD is applied after TF-CM. The student first performs a temporal AR rollout with KV caching. At chunk $i$, the model generates the current clean chunk conditioned on the cached states of previous generated chunks:
\begin{equation}
\tilde{\xv}_{0}^{i}
=
\Gv_{\theta}(\zv^i \mid \mathrm{KV}^{<i}),
\qquad
\mathrm{KV}^{<i}
=
\mathrm{KV}(\tilde{\xv}_{0}^{<i}).
\end{equation}
After $\tilde{\xv}_{0}^{i}$ is generated, it is fed once more into the causal transformer through a cache-update forward pass, which appends its clean-token key/value states to the cache:
\begin{equation}
\mathrm{KV}^{\le i}
=
\mathrm{Append}
\left(
\mathrm{KV}^{<i},
\mathrm{KV}_{\theta^-}(\texttt{sg}[\tilde{\xv}_{0}^{i}])
\right).
\end{equation}

Within each chunk, $\Gv_\theta$ is implemented by few-step self-rollout denoising from pure noise $\zv^i$:
\begin{equation}
\tilde{\xv}^i_{t_N}=\zv^i
\xrightarrow{\ \theta^-\ }
\tilde{\xv}^i_{t_{N-1}}
\xrightarrow{\ \theta^-\ }
\cdots
\xrightarrow{\ \theta^-\ } \tilde{\xv}^i_{t_{1}}
\xrightarrow{\ \theta\ } 
\tilde{\xv}^i_0 ,\quad 0<t_1<t_2<\cdots<t_N=1.
\end{equation}
In each training iteration, the number of simulation steps $N$ is randomly sampled from $[1,N_{\max}]$. Each transition can be instantiated as CM-style reverse denoising followed by forward noising, e.g., under the RF schedule,
\begin{equation}
\tilde{\xv}_{t_{n-1}}^{i}
=
(1-t_{n-1})\fv_{\theta}
\left(
\tilde{\xv}_{t_n}^{i},t_n
\mid
\mathrm{KV}^{<i}
\right)+t_{n-1}\epsilonv_n,\quad\epsilonv_n\sim\Nc(\vect0,\Iv).
\end{equation}
The final output $
\tilde{\xv}_{0}
=
[\tilde{\xv}_{0}^{1},\ldots,\tilde{\xv}_{0}^{N_{\mathrm{chunk}}}]
$ enters the DMD loss. Following standard practice~\citep{yin2024improved,huang2025selfforcing}, we apply gradient truncation to make SF-DMD memory-efficient. The intermediate denoising steps and previous-chunk KV caches are detached (indicated by $\theta^-$). Only the final denoising step $t_1\to0$ of each chunk is kept differentiable (indicated by $\theta$), which the DMD loss is back-propagated through.

\subsubsection{JVP-based Causal Distillation with Teacher-Forcing sCM/MeanFlow}
TF-sCM uses the same packed causal forward as TF and TF-dCM, but replaces the finite-step consistency target with a continuous-time tangent target. The clean context is kept fixed, while the noisy tokens move along the causal teacher ODE. Under the RF schedule, define the causal teacher velocity on the noisy branch as
\begin{equation}
\vv_{\mathrm{teacher}}^{\mathrm{TF}}
=
\left[
\vv_{\mathrm{teacher}}
\left(
[\xv_0^{\mathrm{clean}},\xv_t^{\mathrm{noisy}}],
[\vect 0^{\mathrm{clean}},\vect t^{\mathrm{noisy}}];
\Mv_{\mathrm{TF}}
\right)
\right]_{\mathrm{noisy}} .
\end{equation}
The RF consistency map on the noisy branch is
\begin{equation}
\left[
\fv_\theta^{\mathrm{TF\text{-}RF}}
\right]_{\mathrm{noisy}}
=
\xv_t^{\mathrm{noisy}}
-
t
\left[
\vv_\theta
\left(
[\xv_0^{\mathrm{clean}},\xv_t^{\mathrm{noisy}}],
[\vect 0^{\mathrm{clean}},\vect t^{\mathrm{noisy}}];
\Mv_{\mathrm{TF}}
\right)
\right]_{\mathrm{noisy}} .
\end{equation}
Its continuous-time tangent along the causal teacher trajectory is
\begin{equation}
\label{eq:tf-scm-tangent}
\begin{aligned}
\hv_{\mathrm{TF\text{-}sCM}}
&=
\vv_{\mathrm{teacher}}^{\mathrm{TF}}
-
\left[
\vv_{\theta^-}
\left(
[\xv_0^{\mathrm{clean}},\xv_t^{\mathrm{noisy}}],
[\vect 0^{\mathrm{clean}},\vect t^{\mathrm{noisy}}];
\Mv_{\mathrm{TF}}
\right)
\right]_{\mathrm{noisy}} \\
&\quad
-
t
\left[
\texttt{JVP}
\left(
\vv_{\theta^-},
\left(
[\xv_0^{\mathrm{clean}},\xv_t^{\mathrm{noisy}}],
[\vect 0^{\mathrm{clean}},\vect t^{\mathrm{noisy}}]
\right),
\left(
[\vect 0^{\mathrm{clean}},\vv_{\mathrm{teacher}}^{\mathrm{TF}}],
[\vect 0^{\mathrm{clean}},\vect 1^{\mathrm{noisy}}]
\right);
\Mv_{\mathrm{TF}}
\right)
\right]_{\mathrm{noisy}} .
\end{aligned}
\end{equation}
Here the JVP is computed through the same TF-masked packed forward as the primal prediction. The tangent of the clean context is zero, and only the noisy branch follows the teacher velocity.

The TF-sCM objective is then
\begin{equation}
\label{eq:tf-scm-loss}
\Lc_{\mathrm{TF\text{-}sCM}}(\theta)
=
\E_{\xv_0,\epsilonv,t}
\left[
\left\|
\Delta\vv_\theta^{\mathrm{TF}}
-
\frac{
w(t)\hv_{\mathrm{TF\text{-}sCM}}
}{
w^2(t)\|\hv_{\mathrm{TF\text{-}sCM}}\|_2^2+c
}
\right\|_2^2
\right],
\end{equation}
where
\begin{equation}
\Delta\vv_\theta^{\mathrm{TF}}
=
\left[
\vv_\theta
\left(
[\xv_0^{\mathrm{clean}},\xv_t^{\mathrm{noisy}}],
[\vect 0^{\mathrm{clean}},\vect t^{\mathrm{noisy}}];
\Mv_{\mathrm{TF}}
\right)
-
\vv_{\theta^-}
\left(
[\xv_0^{\mathrm{clean}},\xv_t^{\mathrm{noisy}}],
[\vect 0^{\mathrm{clean}},\vect t^{\mathrm{noisy}}];
\Mv_{\mathrm{TF}}
\right)
\right]_{\mathrm{noisy}} .
\end{equation}

A subtle but important design choice is to use the RF-native form of sCM, rather than wrapping the RF velocity model into TrigFlow and applying the TrigFlow-sCM objective as in rCM~\citep{zheng2025large}. Although different diffusion noise schedules, such as TrigFlow and RF, are analytically convertible up to a time-dependent scaling~\citep{zheng2023improved}, they generally induce different normalized MSE objectives for sCM (Appendix~\ref{app:trigflow-scm-vs-rf-scm}). In the bidirectional setting, rCM finds the TrigFlow wrapper beneficial for stability. However, in our causal TF setting, the TrigFlow-wrapped TF-sCM results in degraded generation quality, whereas the RF-native TF-sCM produces more smooth outputs.

\subsubsection{Extension to Noisy Context and Custom Step Schedule}
\input{figures/diagonal}
Noisy context and custom step schedules~\citep{liu2026diagdistill} are two simplest and most effective inference acceleration techniques for AR video diffusion distillation. Both TF and SF can naturally incorporate them, as illustrated in Fig.~\ref{fig:diagonal}.

\paragraph{Noisy Context} Unlike LLMs, AR video diffusion must maintain a denoising-time-aware KV cache: standard clean-context AR inference requires an additional clean-context encoding pass after the denoising steps of each chunk, so an $N$-step causal diffusion model effectively costs $N+1$ number of function evaluations (NFEs) per chunk. Noisy context removes this extra pass by reusing the KV states from the last denoising step as the context for subsequent chunks, reducing the effective latency from $N+1$ to $N$ NFEs. Besides acceleration, noisy context can improve long-horizon robustness, as residual noise acts as a low-pass filter that suppresses accumulated high-frequency artifacts while preserving coarse motion dynamics~\citep{huang2025live}.

In TF, noisy context is incorporated by replacing the clean history in the packed TF forward with noisy historical tokens at the corresponding context timestep, while the loss remains applied only to the current target block. In SF, noisy context is used directly during AR rollout. Although introducing noisy context in the TF stages would better align with inference, we find it sufficient in practice to apply it only in the final SF stage.
\paragraph{Custom Step Schedule} The number of denoising steps can also vary across chunks. In text-to-video generation, the first chunk is typically more demanding because it establishes the global scene, layout, and appearance, whereas later chunks mainly extend the video conditioned on previous context. We therefore allow a chunk-dependent step schedule
\begin{equation}
[N_1,N_2,\ldots,N_{N_{\mathrm{chunk}}}],
\end{equation}
where $N_i$ denotes the number of denoising steps for chunk $i$. For example, a nominal 2-step model can use $[4,2,2,\ldots]$, allocating extra computation only to the first chunk.

For SF-DMD training, we cycle the rollout length by the training iteration. For example, for a target schedule $[4,2,2,\ldots]$, SF-DMD repeatedly cycles through
$
[1,1,1,\ldots]
\rightarrow
[2,2,2,\ldots]
\rightarrow
[3,2,2,\ldots]
\rightarrow
[4,2,2,\ldots]
\rightarrow
[1,1,1,\ldots]
\rightarrow
\cdots$. This cycling strategy is important because SF-DMD only back-propagates through the final denoising step of each chunk. Cycling the rollout length makes different denoising intervals appear as the final differentiable step across iterations, rather than supervising only the last interval of the maximum-step sampler.

\subsection{Infrastructure}
\input{tables/infra}
Causal-rCM is designed as an algorithm-infrastructure recipe. Its main infrastructure goal is to make causal training paradigms (TF, DF, and SF), continuous-time JVP-based CMs, and large-scale parallel training mutually compatible. Achieving this requires careful co-design of attention-mask specification, KV caching, FSDP2, context parallelism, activation checkpointing, FlashAttention-2 JVP kernels, and replayed back-propagation. Table~\ref{tab:wan_causal_distillation_infra} summarizes the resulting system-level coverage and highlights the infrastructure advantages of Causal-rCM over other widely used codebases.

\subsubsection{Main Components}
\paragraph{FlashAttention-2 JVP Kernel with Custom Masks}

Continuous-time CMs require the tangent of the network output along the teacher ODE. Computing this tangent with a generic \texttt{torch.func.jvp} over unfused attention is impractical for large video transformers due to the materialization of large attention intermediates and the resulting memory overhead. To enable JVP through fused attention under TF masks, we build on the FlashAttention2-JVP kernel in rCM~\citep{zheng2025large} and extend it to support custom masks. The TF mask is represented as admissible query-key ranges rather than materialized dense matrices. The details are presented in Appendix~\ref{app:fa2-jvp-custom-mask}.

\paragraph{Parallelisms}

We use FSDP2~\citep{zhao2023pytorch} as a ZeRO-3-style sharding backend: parameters, gradients, and optimizer states are partitioned across data-parallel ranks, and each module materializes full parameters only for its local computation. This reduces per-GPU model-state memory and makes it feasible to train large video DiTs with student, teacher, fake-score, and EMA networks in the same distillation pipeline. We use distributed checkpointing (DCP) to save and restore the sharded model and optimizer states directly across ranks, avoiding the need to gather full model states on a single process.

We use flattened Ulysses-style context parallelism (CP)~\citep{jacobs2023deepspeed} to shard the long video-token sequence across ranks. Specifically, the spatiotemporal video tokens are first flattened into a single sequence, and CP partitions this flattened sequence dimension across \texttt{P} devices. Before attention, each GPU holds a shard of size \texttt{[B, H, L/P, C]} for QKV. An all-to-all operation then redistributes QKV to \texttt{[B, H/P, L, C]} for local attention, followed by another all-to-all to restore the sequence partition of the attention output $\Ov$. A key design choice is to make CP transparent to the outer algorithm: the network interface always takes and returns the global full sequence, independent of CP size, while the network internally handles local sequence shards, all-to-all attention, and output gathering. 

\paragraph{Activation Checkpointing}

We use selective activation checkpointing (SAC) to reduce activation memory by recomputing only selected parts of the network during backward. Unlike vanilla region-based \texttt{torch.utils.checkpoint}, SAC provides finer-grained control over which operations are recomputed and which intermediates are preserved. In practice, we apply SAC mainly to compute-heavy stateless regions such as attention and MLP blocks, while leaving lightweight or stateful operations outside checkpointed regions.

\paragraph{KV Cache}

The KV cache is used by causal rollout execution and inference. We distinguish three cache modes: \texttt{disabled} mode for ordinary packed training, \texttt{append} mode for committing a generated chunk into the cache, and \texttt{readonly} mode for generating the current chunk while attending to previously committed chunks. Cached K/V tensors are detached by construction, which prevents gradients from propagating through previous chunks and keeps SF-DMD memory bounded. The cache also records chunk boundaries, so a readonly forward can expose only the prefix needed by the current block. This supports both standard AR rollout and variants such as noisy context, where the final denoising forward can be reused as the context state.

We support both pre-RoPE and post-RoPE key caching. Post-RoPE caching is simple and efficient because cached keys can be reused directly. Pre-RoPE caching is useful when the same cached content may need different positional treatment, e.g., for length extrapolation or alternative position indexing~\citep{yesiltepe2026infinity,yi2025deep,li2026rolling,kim2026memrope}. The implementation keeps this choice inside the attention context so that the high-level rollout code does not need to distinguish the two cases.

\paragraph{Replayed Back-propagation}

SF-DMD generates on-policy videos through AR rollout. In the standard execution with gradient-truncation, the final differentiable denoising steps of all chunk are kept in the computational graph, which can be memory-intensive for long videos. We therefore provide an optional replayed back-propagation mode~\citep{hong2025relic} as a memory-saving implementation. The rollout is first constructed without gradients, while storing the final noisy input, timestep, detached KV cache, and DMD target for each chunk. Then, each chunk's final denoising step is recomputed with gradients enabled, and its gradient is back-propagated separately with gradient accumulation. This trades additional computation for lower activation memory. We deliberately reserve this replayed path for SF-DMD: TF, DF, and TF-CM remain packed, since replaying differentiable prefix-KV computation offers limited additional benefit once SAC is enabled.

\subsubsection{Compatibility Design}

A major goal of Causal-rCM is to make advanced causal training features composable. In practice, many components that work independently can conflict when used together. We therefore implement compatibility at the level of execution semantics rather than as independent feature switches.

\paragraph{SAC \(\times\) FlexAttention.}
Packed TF/DF/TF-CM training relies on custom-mask attention. In the FlexAttention path, the attention pattern is specified by a \texttt{mask\_mod} function and lowered by the PyTorch compiler into a specialized fused attention kernel. To make this compatible with SAC, we use \texttt{torch>=2.10} together with
\[ \texttt{torch.\_inductor.config.wrap\_inductor\_compiled\_regions = True} \]
which exposes Inductor-compiled FlexAttention calls to SAC as explicit checkpointable regions, internally represented as \texttt{inductor\_compiled\_code}.

\paragraph{SAC \(\times\) self-forcing.}
SF-DMD rollout is stateful because KV caches and causal metadata evolve across chunks. We make this compatible with SAC by separating persistent cache storage from per-forward causal state: historical K/V tensors are stored as detached context, while each forward constructs a fresh \texttt{CausalInferenceState} describing the current chunk, cache range, and append/read-only mode for future recomputation. The inference state is not reused through in-place updates, so checkpoint recomputation reconstructs the same causal context as the original forward. Cache-append forwards are kept outside checkpointed execution, so recomputation never replays cache mutation; checkpointed regions only read a fixed causal context.

\paragraph{JVP \(\times\) FSDP2.}
Following rCM, we implement JVP at the layer level, rather than applying a global \texttt{torch.func.jvp} to an FSDP2-wrapped model. Each layer exposes a paired primal-tangent interface, taking \((\xv,\tv\xv)\) as input and returning \((\yv,\tv\yv)\). This corresponds to an \texttt{FSDP2(JVP)} design instead of \texttt{JVP(FSDP2)}. FSDP2 continues to manage parameter materialization, sharding, and gradient reduction at layer boundaries, while tangent propagation is performed locally within each layer's forward computation.

\paragraph{JVP \(\times\) Ulysses CP.}
Ulysses CP extends naturally to JVP because tangent tensors follow the same communication pattern as their primal counterparts. Specifically, $\vtQ,\vtK,\vtV$ are all-to-all exchanged together with $\vQ,\vK,\vV$, the local attention computation is replaced by our custom-mask FlashAttention-2 JVP kernel, and the resulting $\vtO$ is returned through the same output all-to-all as $\vO$. We reuse the JVP-compatible distributed-attention design from rCM, while adding custom-mask support for packed TF/DF/TF-CM training.

\paragraph{KV cache \(\times\) Ulysses CP.}
For rollout execution, cached K/V tensors must be compatible with Ulysses CP. We use a post-all-to-all KV cache, where the cache is stored in the same \texttt{[B, H/P, L, C]} layout as exposed to local attention. Each CP rank directly reuses its head-sharded, full-sequence cached K/V states. This avoids repeatedly converting old cache entries between global and CP-local layouts.

\section{Experiments}
\subsection{Setup}
\input{tables-exp/exp-detail}
\paragraph{Models and Datasets.}
We conduct the main streaming video generation experiments on Wan2.1 T2V~\citep{wan2025} at 480p resolution. Videos are generated at $832\times480$ spatial resolution with 81 RGB frames, corresponding to 21 latent frames after VAE temporal compression. Training uses the synthetic T2V data provided by rCM~\citep{zheng2025large}, generated by the bidirectional Wan2.1-14B teacher with 100-step Euler sampling, shift 3.0, and CFG scale 5.0. We use Wan2.1-1.3B as the main student model and use Wan2.1-14B teachers for distillation.

We evaluate two causal chunk patterns. The \textit{frame-wise} setting, denoted by \texttt{c1-1}, uses one initial latent frame and then one-latent-frame chunks. The \textit{chunk-wise} setting, denoted by \texttt{c3-3}, uses one initial latent chunk of three frames and then three-latent-frame chunks. The same chunk pattern is used consistently for packed TF/DF/TF-CM masks, SF-DMD rollout, KV-cache inference, and streaming evaluation.

\paragraph{Training.}
Causal-rCM uses a three-stage training recipe. We report the main hyperparameters in Table~\ref{tab:training-config}. For TF-CM, we use 14B causal teachers trained with TF. For SF-DMD, we use 14B birectional teacher and fake score networks.

For few-step SF-DMD, we use RF sampling schedules with a maximum of 4 denoising steps. The 4-step sampler uses intermediate times $[15/16,5/6,5/8]$. The 2-step sampler uses 4 steps for the first chunk and 2 steps for later chunks, with schedule $[[15/16,5/6,5/8],[5/6]]$. The 2-step noisy-context variant uses schedule $[[15/16,5/6,5/8],[5/8]]$ and reuses the final denoising forward as the context cache. The 1-step variant uses 4 steps for the first chunk and 1 step for later chunks, with schedule $[[15/16,5/6,5/8],[]]$.

\paragraph{Evaluation Metrics.}
For streaming quality, we evaluate text-to-video generation with VBench-T2V~\citep{huang2024vbench}, reporting the total score as well as the quality and semantic sub-scores.

For inference efficiency, we report the number of function evaluations (NFE), throughput in frames per second (FPS), first-chunk latency, and second-chunk latency. All efficiency measurements are conducted with batch size 1 on a single H100 GPU. The reported FPS and latency include \textit{both diffusion sampling and VAE decoding}.

\subsection{Results}
\subsubsection{Streaming Video Generation}
\paragraph{Main Results.}
Table~\ref{tab:main_results} compares Causal-rCM against bidirectional Wan2.1 and streaming video generation baselines, including Self-Forcing~\citep{huang2025selfforcing}, LongLive~\citep{longlive}, Causal Forcing~\citep{zhu2026causal}, and AnyFlow~\citep{gu2026anyflow}. We report both frame-wise and chunk-wise results. Causal-rCM achieves state-of-the-art streaming quality while supporting 4-step, 2-step, 2-step noisy-context, and 1-step inference schedules.
\input{tables-exp/main}

\paragraph{Performance under Custom Step Schedule and Noisy Context.}
Table~\ref{tab:main_results} shows an interesting behavior under custom step schedules. In the frame-wise setting, the 1-step and 2-step Causal-rCM models outperform the 4-step variant, which is counter-intuitive at first glance. We attribute this to the nature of the frame-wise setting: each AR chunk contains only a single latent frame and therefore has no internal temporal structure to denoise. In this case, allocating many denoising steps to every future chunk can over-emphasize autoregressive feedback errors, especially considering the gradient truncating strategy of SF-DMD. Empirically, we observe that 4-step frame-wise SF-DMD is more prone to camera drift, e.g., a consistent leftward camera rotation across samples, and can only be trained stably for about 1k iterations. In contrast, using 1 or 2 steps for later chunks largely suppresses this drift and allows stable training for around 3k iterations. Since each future chunk contains only one latent frame, 1--2 denoising steps are already sufficient to generate the frame, and the reduced rollout depth improves stability.

The trend is different in the chunk-wise setting, where each chunk contains three latent frames and therefore has non-trivial internal temporal correlation. Here, a deeper 4-step sampler provides a better denoising trajectory for modeling motion and intra-chunk consistency, leading to the best overall score. This suggests that the optimal step schedule depends on the temporal span of each AR chunk: frame-wise generation benefits more from shallow, stable rollout, while chunk-wise generation benefits from additional denoising depth.

Noisy context further improves inference efficiency by eliminating the extra clean-context KV encoding pass, reducing the effective cost from $N+1$ to $N$ NFEs per chunk. Comparing 2-step sampling with noisy context against 1-step sampling, we find that 1-step sampling is better in the frame-wise setting, while 2-step sampling with noisy context is better in the chunk-wise setting. This is consistent with the above observation. For single-frame chunks, the extra denoising step brings limited benefit, while the residual noise in the context can directly affect fine-grained details in frame-level prediction. For three-frame chunks, the chunk contains a higher-dimensional and more redundant spatiotemporal token group. In this regime, Gaussian perturbations are less likely to destroy the entire chunk-level structure uniformly, and much of the motion and coarse semantic context can still be preserved~\citep{hoogeboom2023simple}. Therefore, 2-step sampling with noisy context can retain the benefit of an additional denoising step for intra-chunk temporal coherence.

\paragraph{Comparison between TF-dCM and TF-sCM.}

\input{figures/dcm_vs_scm}

Fig.~\ref{fig:dcm-vs-scm} compares TF-dCM and TF-sCM before the final SF-DMD stage. TF-sCM consistently provides a stronger initialization with over $10\times$ fewer training iterations. In the frame-wise setting, TF-sCM reaches above 81.8 VBench-T2V score within 1-2k iterations, already surpassing TF-dCM trained for 10k iterations. The gap is even clearer in the chunk-wise setting, where TF-sCM reaches above 83 within 1-2k iterations, while TF-dCM improves more slowly and remains lower after much longer training.

\paragraph{Ablation Studies on Initialization Strategies.}
\input{tables-exp/ablation}
\input{figures/vbench_curve}
Table~\ref{tab:ablation} ablates the initialization strategies of SF-DMD. We compare causal diffusion initializations from DF and TF, ODE-pair knowledge distillation variants (DF-KD and TF-KD), and teacher-forcing consistency initializations (TF-dCM and TF-sCM). The corresponding training curves are shown in Fig.~\ref{fig:vbench-curve}.

\input{figures/oversmooth}

In the frame-wise setting, TF-CM initialization achieves the best overall performance, with DF and TF-KD also providing competitive alternatives. Although TF-sCM starts from a stronger initial model, TF-dCM is more stable during SF-DMD and supports longer refinement, leading to a higher peak score. In the chunk-wise setting, DF/TF initialization achieves the highest VBench-T2V scores, close to 85. However, as shown in Fig.~\ref{fig:oversmooth}, these models often produce over-smoothed and over-saturated textures, such as water, hair, and leaves, with noticeably fewer fine-grained details. Considering both VBench scores and qualitative inspection, TF-CM initialization is still the most reliable choice. Among the two TF-CM variants, TF-sCM slightly outperforms TF-dCM while requiring fewer SF-DMD iterations.

\subsubsection{Interactive World Model}
\input{figures/interactive_overview}

We further apply Causal-rCM to Cosmos 3~\citep{nvidia2026cosmos3}, an omnimodal world model based on a two-tower Mixture-of-Transformers architecture. Cosmos 3 separates an understanding tower (UND) for text and prompt reasoning from a generation tower (GEN) for vision, action, and sound tokens, while sharing the multimodal attention layers and unified 3D mRoPE across modalities. In the original generator mode, GEN tokens use bidirectional self-attention for multimodal denoising. To support interactive world modeling, we convert the GEN vision stream into a temporal-causal autoregressive diffusion stack (Fig.~\ref{fig:interactive-overview}).

We treat each latent video frame as a vision supertoken, which contains all spatial latent tokens of that frame. Temporal-causal attention is applied at the supertoken level: future vision supertokens are masked from past and current ones, while spatial tokens within the same vision supertoken remain fully bidirectional.

The same causal stack supports text-to-video, image-to-video, and forward-dynamics (action-conditioned) modeling. In text-to-video, all vision supertokens are generated from text conditioning. In image-to-video and forward dynamics, the first vision supertoken is provided as clean context, and the model predicts future vision supertokens autoregressively. For forward dynamics, action supertokens are treated as input conditions. A null action supertoken is used for the first frame, and real action supertokens are aligned by unified 3D mRoPE to the next generated vision supertoken, so that action \(A_i\) controls the transition from state \(V_i\) to \(V_{i+1}\).

\input{figures/interactive_av}

As shown in Fig.~\ref{fig:interactive-av}, the interactive Cosmos 3 model supports streaming control: given the same initial scene, the generated future frames follow distinct trajectories under left-turn, right-turn, and stay-forward controls.

\section{Related Work}
\paragraph{Differential information and JVPs in generative modeling.}
Differential information has played an important role in diffusion ODEs beyond standard first-order denoising supervision.
High-order denoising score matching shows that first-order score matching is insufficient for maximum-likelihood diffusion ODE training, and controls higher-order score errors to tighten the likelihood gap~\citep{lu2022maximum}.
Subsequent work improves diffusion ODE likelihood estimation and training with velocity parameterization, variance reduction, and high-order flow-matching objectives~\citep{zheng2023improved}.
DPM-Solver-v3 further uses empirical model statistics of a pretrained diffusion model to derive improved ODE solver coefficients, and also reveals numerical issues related to time derivatives in diffusion networks~\citep{zheng2023dpm}.
More recently, sCM, MeanFlow, AYF, and FACM use JVPs as a direct training signal for continuous-time consistency or flow-map objectives~\citep{lu2024simplifying,geng2025mean,sabour2025align,peng2025facm}.
rCM scales JVP-based consistency distillation to large image and video diffusion models by making JVP computation compatible with FlashAttention, FSDP, and context parallelism, and combines it with DMD regularization~\citep{zheng2025large}.
Causal-rCM extends this line to autoregressive video diffusion, applying JVP-based teacher-forcing sCM under clean causal contexts as a structured initialization for self-forcing DMD.

\paragraph{Forward-reverse complementarity in distillation objectives.}
A growing set of few-step methods can be viewed as combining a coverage-preserving forward component with a quality- or reward-seeking reverse component. 
For text-to-image generation, recent practical studies standardize large-scale few-step distillation recipes for strong text-conditioned teachers, and empirically compare sCM with MeanFlow~\citep{pu2025few}. Flow-map methods distill teacher ODE behavior more directly. FreeFlow~\citep{tong2025flow} performs data-free flow-map distillation by sampling from the prior and querying teacher dynamics on student-induced flow-map states, with an additional correction objective to mitigate compounding errors. In contrast, \(\pi\)-Flow~\citep{chen2025pi} is more explicitly on-policy by matching teacher velocities along the student policy's own ODE trajectory.
Distribution-matching methods improve few-step quality but may sacrifice diversity; recent variants therefore introduce role separation, RL signals, or adversarial flow objectives to balance mode coverage and mode seeking~\citep{jiang2025distribution,wu2026diversity,cheng2025twinflow,lin2026continuous}. 
This complementarity is especially explicit in recent long-video work: \citet{cai2026mode} pair a supervised global flow-matching head for long-range structure with a local DMD head for short-window fidelity, while HiAR~\citep{zou2026hiar} observes that self-rollout reverse-KL distillation can amplify low-motion shortcuts and adds a forward-KL regularizer to preserve motion diversity.

\paragraph{Video and autoregressive diffusion distillation.}
Video distillation must handle temporal consistency and long-horizon error accumulation in addition to per-frame visual quality. 
Self-Forcing~\citep{huang2025selfforcing} and APT2~\citep{lin2025autoregressive} are representative works that propose self-forcing as an on-policy distillation paradigm for mitigating exposure bias in AR generation. In particular, APT2 initializes self-forcing with teacher-forcing consistency distillation, but relies on a relatively cumbersome GAN objective during the self-forcing stage. Concurrent to our work, Causal Forcing++~\citep{zhao2026causal} also combines teacher-forcing consistency with self-forcing DMD, while we implement JVP-based continuous-time consistency under teacher forcing, and provide a systematic algorithm-and-infrastructure open recipe with holistic evaluation. Apart from the CM route, Transition Matching Distillation matches multi-step video denoising trajectories with few-step transition processes using conditional flow heads, followed by distribution matching on flow-head rollouts~\citep{nie2026transition}. 
AnyFlow shifts video distillation from endpoint consistency to arbitrary-interval flow-map transitions and uses backward simulation for on-policy distillation in both bidirectional and causal architectures~\citep{gu2026anyflow}. 
Other recent work studies from-scratch few-step video training with efficient solution-flow objectives~\citep{park2026eflow}, video-specific distillation losses for oversaturation and temporal collapse~\citep{you2026adaptive}. Adversarial refinement has also been explored for one-step AR video generation, e.g., by augmenting DMD with a noised-latent GAN loss~\citep{feng2026one} or by using asymmetric adversarial distillation after distribution-matching warm-up~\citep{li2026aad}. The distilled models are orthogonal to attention-level acceleration and could be further combined with sparse attention techniques~\citep{zhang2025sla,zhang2026sla2}, as demonstrated by TurboDiffusion~\citep{zhang2025turbodiffusion}, which combines rCM with attention acceleration and quantization.

\section{Limitations and Future}
\label{sec:limitations-future}

\paragraph{Limitations.}
Although Causal-rCM provides an effective algorithm-infrastructure recipe for autoregressive diffusion distillation, several limitations remain. First, frame-wise T2V training with long rollout depth is still fragile. In this setting, the 4-step SF-DMD model tends to develop camera drift after extended training, e.g., a consistent directional camera bias, and therefore cannot be trained for a long duration. This issue could be eliminated in action-conditioned interactive settings, where actions provide an explicit motion prior and reduce the ambiguity of camera evolution. Second, the best initialization before SF-DMD does not always translate into the best final model. TF-sCM gives a stronger pre-SF-DMD initialization than TF-dCM, but in the frame-wise setting, TF-dCM can be more stable under long SF-DMD refinement and achieve a higher final peak. This suggests that initialization quality and refinement stability are not fully aligned. Third, fully joint optimization like rCM remains challenging. In our causal setting, joint training tends to lower the VBench ceiling, so we currently use a staged pipeline. This could be attributed to the distribution gap between the causal teacher and the bidirectional teacher. Finally, the current custom-mask FlashAttention JVP kernel is implemented in Triton. As a result, the per-iteration speed of TF-sCM is only comparable to TF-dCM with standard FlashAttention-2, lacking behind more advanced kernels like FlashAttention-3/4.

\paragraph{Future directions.}
A natural next step is to make the staged recipe more systematic. Table~\ref{tab:distillation_pipeline_insight} summarizes our high-level view: current distillation methods can be interpreted as subsets of two ultimate pipelines, a CM route and a CTM route, each with bidirectional and causal variants. Discrete-time methods (dCM, MeanFlow with finite difference estimation) could be the warmup stage for continuous-time JVP ones (sCM, MeanFlow) to enhance stability.

\input{tables/distillation_pipeline}

Beyond algorithmic design, future work should improve the underlying systems stack. Better kernels for custom attention, JVP, and KV-cache execution, together with runtime features such as \texttt{torch.compile}, CUDA Graphs, and \texttt{NVFP4} could further reduce overhead and make large-scale training and inference more efficient.

\clearpage
\setcitestyle{numbers}
\bibliographystyle{plainnat}
\bibliography{main}

\newpage
\setcitestyle{authoryear,round}
\appendix
\input{sections/appendix_trigflow_scm_vs_rf_scm}
\input{sections/appendix_fa2_jvp_custom_mask}
\end{document}

%% file: packages.tex
\usepackage[authoryear,round]{natbib}

\usepackage[utf8]{inputenc} 
\usepackage[T1]{fontenc}    

\usepackage{parskip}        
\usepackage{url}            
\usepackage{booktabs}       
\usepackage{amsfonts}       
\usepackage{nicefrac}       
\usepackage{microtype}      
\usepackage{xcolor}         
\usepackage[dvipsnames]{xcolor} 
\usepackage{graphicx}
\usepackage{animate}        
\usepackage{subcaption}
\usepackage{tabularx}
\usepackage{makecell}
\usepackage{adjustbox}
\usepackage{setspace}
\newcolumntype{M}[1]{>{\centering\arraybackslash}m{#1}}
\usepackage{float}
\usepackage{tikz}
\usetikzlibrary{positioning,shapes,arrows}
\usepackage{amsmath,amsfonts,bm, bbm,leftindex}
\usepackage{multirow}
\usepackage{comment}
\usepackage{gensymb}
\usepackage{lipsum}
\usetikzlibrary{arrows.meta, positioning, fit}
\usepackage[para]{threeparttable}
\usepackage{tikz}
\usetikzlibrary{tikzmark}

\usepackage[most]{tcolorbox}
\usepackage{fancyvrb}
\usepackage{fvextra}
\usepackage{dashrule}
\usepackage{nicematrix}

\usepackage{multicol}



%% file: common.tex
\def\eqref#1{equation~\ref{#1}}

\def\1{\bm{1}}

\DeclarePairedDelimiterX{\infdivx}[2]{(}{)}{%
	#1\;\delimsize\|\;#2%
}

\newcommand{\kl}{D_{\mathrm{KL}}\infdivx}

\newcommand{\vect}[1]{\bm{#1}}

\newcommand{\x}{\xv}
\newcommand{\y}{\yv}

\newcommand{\dm}{\mathrm{d}}

\newcommand{\softmax}{\mathrm{softmax}}

\newcommand{\E}{\mathbb{E}}

\newcommand{\epsilonv}{\vect\epsilon}

\newcommand{\fv}{\vect f}
\newcommand{\gv}{\vect g}
\newcommand{\hv}{\vect h}

\newcommand{\tv}{\vect t}

\newcommand{\vv}{\vect v}

\newcommand{\xv}{\vect x}
\newcommand{\yv}{\vect y}
\newcommand{\zv}{\vect z}

\newcommand{\Fv}{\vect F}
\newcommand{\Gv}{\vect G}

\newcommand{\Iv}{\vect I}

\newcommand{\Mv}{\vect M}

\newcommand{\Ov}{\vect O}

\newcommand{\Lc}{\mathcal L}

\newcommand{\Nc}{\mathcal N}

\newcommand{\diag}{\mbox{diag}}

\newcommand{\vQ}{\mathbf{Q}}
\newcommand{\vK}{\mathbf{K}}
\newcommand{\vV}{\mathbf{V}}

\newcommand{\vS}{\mathbf{S}}

\newcommand{\vP}{\mathbf{P}}

\newcommand{\vO}{\mathbf{O}}

\newcommand{\vtQ}{\mathbf{tQ}}
\newcommand{\vtK}{\mathbf{tK}}
\newcommand{\vtV}{\mathbf{tV}}
\newcommand{\vtS}{\mathbf{tS}}

\newcommand{\vtO}{\mathbf{tO}}

\newcommand{\vA}{\mathbf{A}}
\newcommand{\vB}{\mathbf{B}}

\newcommand{\vH}{\mathbf{H}}

\makeatletter
\let\save@mathaccent\mathaccent
\newcommand*\if@single[3]{%
  \setbox0\hbox{${\mathaccent"0362{#1}}^H$}%
  \setbox2\hbox{${\mathaccent"0362{\kern0pt#1}}^H$}%
  \ifdim\ht0=\ht2 #3\else #2\fi
  }
\newcommand*\rel@kern[1]{\kern#1\dimexpr\macc@kerna}
\newcommand*\widebar[1]{\@ifnextchar^{{\wide@bar{#1}{0}}}{\wide@bar{#1}{1}}}
\newcommand*\wide@bar[2]{\if@single{#1}{\wide@bar@{#1}{#2}{1}}{\wide@bar@{#1}{#2}{2}}}
\newcommand*\wide@bar@[3]{%
  \begingroup
  \def\mathaccent##1##2{%
    \let\mathaccent\save@mathaccent
    \if#32 \let\macc@nucleus\first@char \fi
    \setbox\z@\hbox{$\macc@style{\macc@nucleus}_{}$}%
    \setbox\tw@\hbox{$\macc@style{\macc@nucleus}{}_{}$}%
    \dimen@\wd\tw@
    \advance\dimen@-\wd\z@
    \divide\dimen@ 3
    \@tempdima\wd\tw@
    \advance\@tempdima-\scriptspace
    \divide\@tempdima 10
    \advance\dimen@-\@tempdima
    \ifdim\dimen@>\z@ \dimen@0pt\fi
    \rel@kern{0.6}\kern-\dimen@
    \if#31
      \overline{\rel@kern{-0.6}\kern\dimen@\macc@nucleus\rel@kern{0.4}\kern\dimen@}%
      \advance\dimen@0.4\dimexpr\macc@kerna
      \let\final@kern#2%
      \ifdim\dimen@<\z@ \let\final@kern1\fi
      \if\final@kern1 \kern-\dimen@\fi
    \else
      \overline{\rel@kern{-0.6}\kern\dimen@#1}%
    \fi
  }%
  \macc@depth\@ne
  \let\math@bgroup\@empty \let\math@egroup\macc@set@skewchar
  \mathsurround\z@ \frozen@everymath{\mathgroup\macc@group\relax}%
  \macc@set@skewchar\relax
  \let\mathaccentV\macc@nested@a
  \if#31
    \macc@nested@a\relax111{#1}%
  \else
    \def\gobble@till@marker##1\endmarker{}%
    \futurelet\first@char\gobble@till@marker#1\endmarker
    \ifcat\noexpand\first@char A\else
      \def\first@char{}%
    \fi
    \macc@nested@a\relax111{\first@char}%
  \fi
  \endgroup
}
\makeatother

\usepackage{algorithm}
\usepackage{algorithmic}

\usepackage{amsthm}

\theoremstyle{plain}

\theoremstyle{definition}

\theoremstyle{remark}

%% file: figures/vbench.tex
\begin{figure}[htbp]
    \centering
    \includegraphics[width=\textwidth]{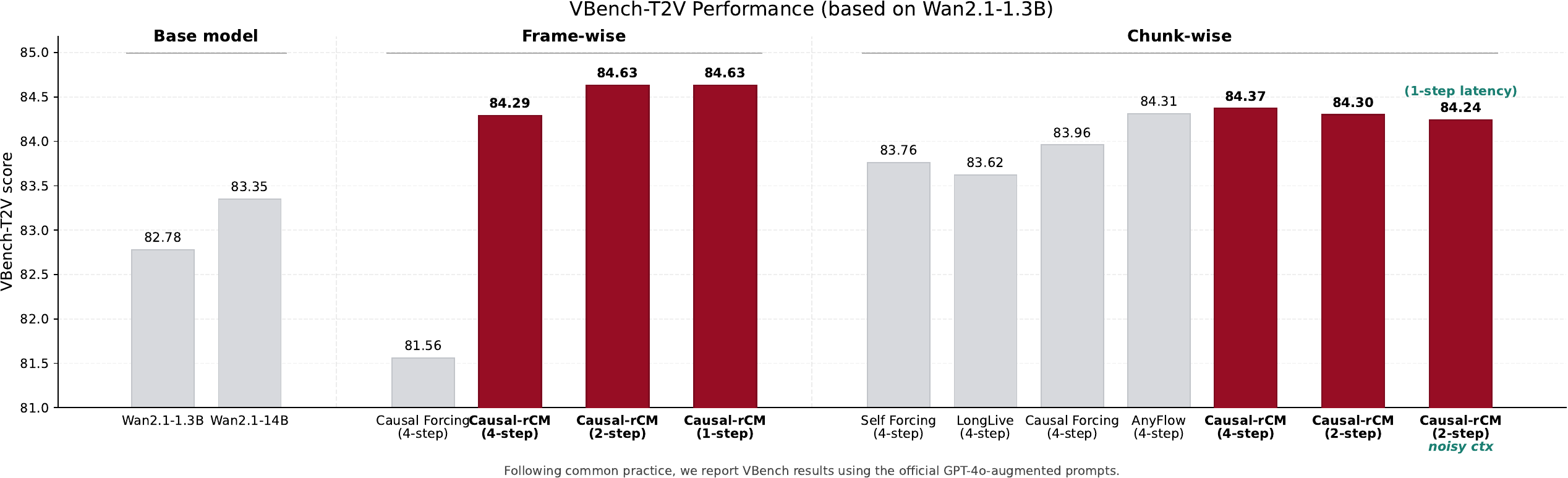}
    \caption{\textbf{State-of-the-art performance of Causal-rCM for streaming video generation (\textcolor{blue}{1-step: 84.63}).} Causal-rCM achieves leading VBench-T2V scores across 1-step, 2-step, and 4-step generation, under both frame-wise and chunk-wise autoregressive regimes.}
    \label{fig:vbench}
\end{figure}

%% file: figures/overview.tex
\begin{figure}[htbp]
    \centering
    \includegraphics[width=\textwidth]{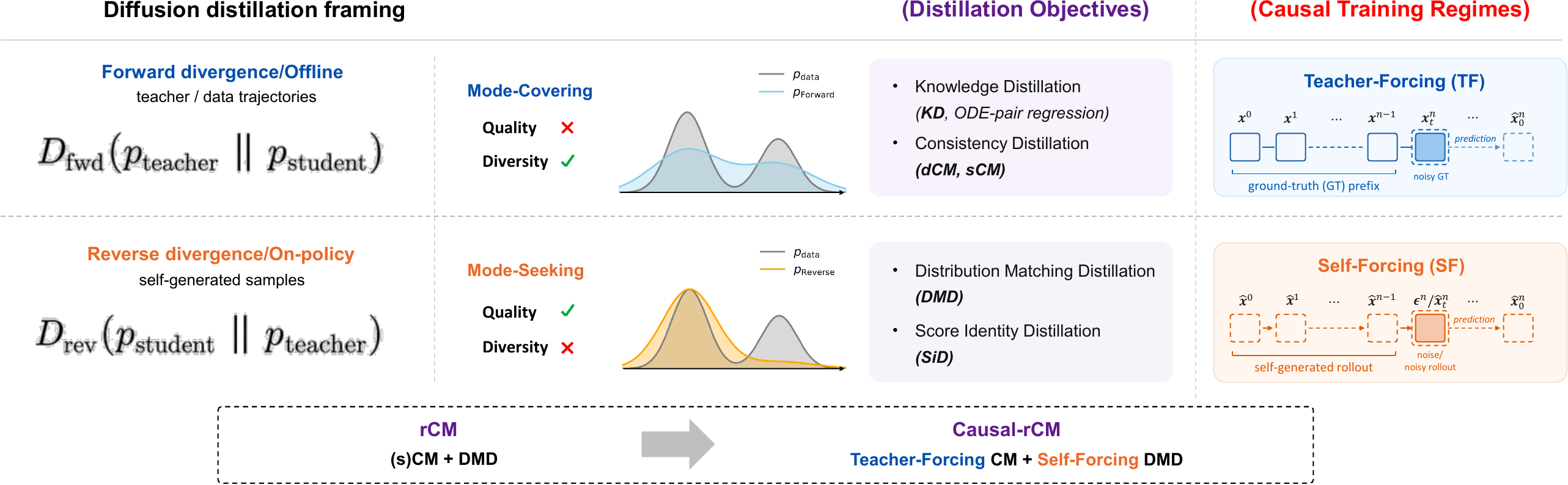}
    \caption{\centering\textbf{A unified divergence perspective of rCM~\citep{zheng2025large} and Causal-rCM.}}
    \label{fig:overall}
\end{figure}

%% file: tables/comparison.tex
\begin{table*}[t]
\centering
\small
\setlength{\tabcolsep}{3.6pt}
\renewcommand{\arraystretch}{1.30}

\caption{\textbf{Forward--reverse objective complementarity across diffusion mid-training, distillation, and RL.}}
\label{tab:forward_reverse_complementarity}

\begin{tabularx}{\textwidth}{
@{}l
>{\raggedright\arraybackslash}p{2.35cm}
>{\raggedright\arraybackslash}X
>{\raggedright\arraybackslash}X
>{\raggedright\arraybackslash}X@{}}
\toprule
Method
& Domain
& Forward Component \newline {\footnotesize (Pretrain / Offline)}
& Reverse Component \newline {\footnotesize (Posttrain / On-policy)}
& Effect / Takeaway \\
\midrule

DDO~\citep{zheng2025direct}
& diffusion / AR mid-training
& diffusion loss on real data
& anti-likelihood diffusion loss on self-generated negatives
& new record FIDs on ImageNet without auxiliary‌ data/model\\

DiffusionNFT~\citep{zheng2025diffusionnft}
& diffusion RL
& forward-process diffusion objective
& reward-ranked positive / negative generated samples
& 25$\times$ efficiency \\

DDRL~\citep{ye2025data}
& diffusion RL
& forward-KL / diffusion-loss regularization to offline data
& GRPO-style reward optimization on generated rollouts
& alleviating reward hacking and diversity collapse \\

rCM~\citep{zheng2025large}
& diffusion distillation
& (s)CM loss on data / teacher trajectories
& DMD loss on student-generated samples
& alleviating mode collapse \\

\textbf{Causal-rCM}
& AR diffusion distillation
& teacher-forcing CM on offline causal contexts
& self-forcing DMD on autoregressive student rollouts
& TF-CM initializes SF with causal structure and mode coverage\\

\bottomrule
\end{tabularx}

\vspace{0.25em}
\begin{minipage}{0.98\textwidth}
\footnotesize
\textit{Notes.}
The complementarity can be realized either in \textit{a single joint stage} or in \textit{a forward-to-reverse order across separate stages}.
We use ``on-policy'' to emphasize self-generated samples or rollouts; in diffusion RL, such data can be online but off-policy in the strict RL sense.
\end{minipage}

\end{table*}

%% file: figures/background.tex
\begin{figure}[htbp]
    \centering
    \includegraphics[width=\textwidth]{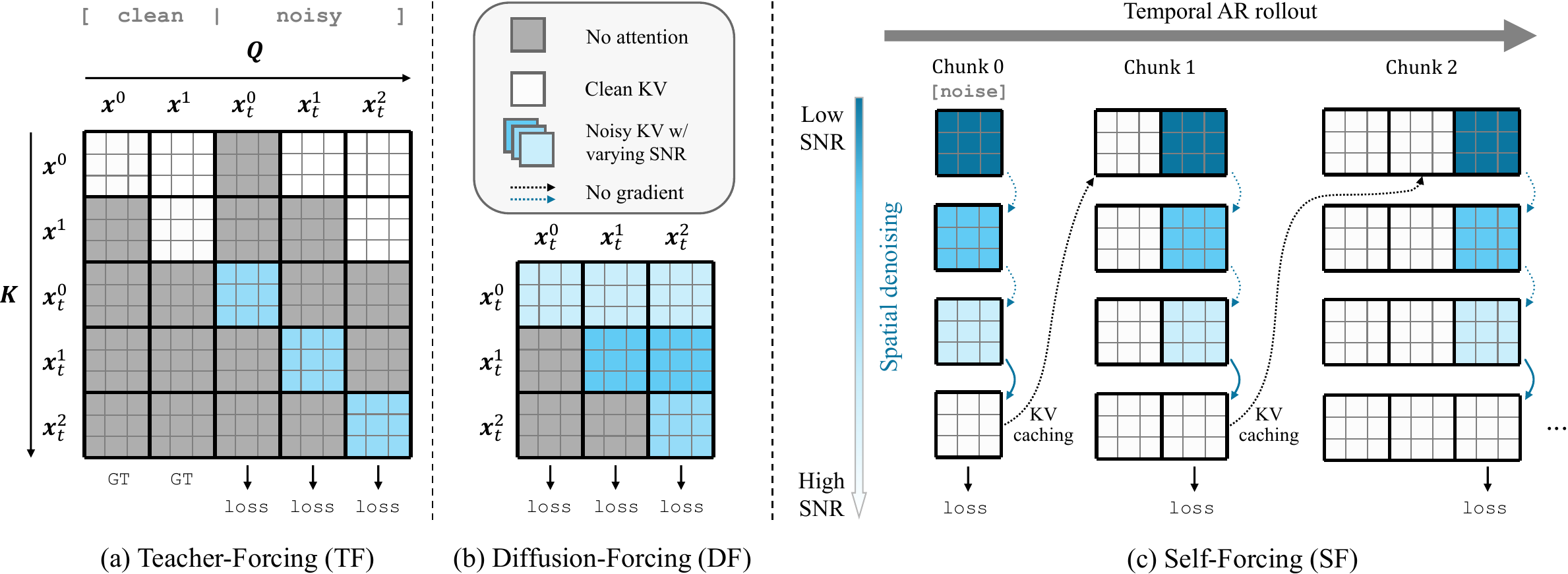}
    \caption{\centering\textbf{Illustration of causal training paradigms, adapted from Self-Forcing~\citep{huang2025selfforcing}.}}
    \label{fig:background}
\end{figure}

%% file: figures/comparison.tex
\begin{figure}[htbp]
    \centering
    \includegraphics[width=\textwidth]{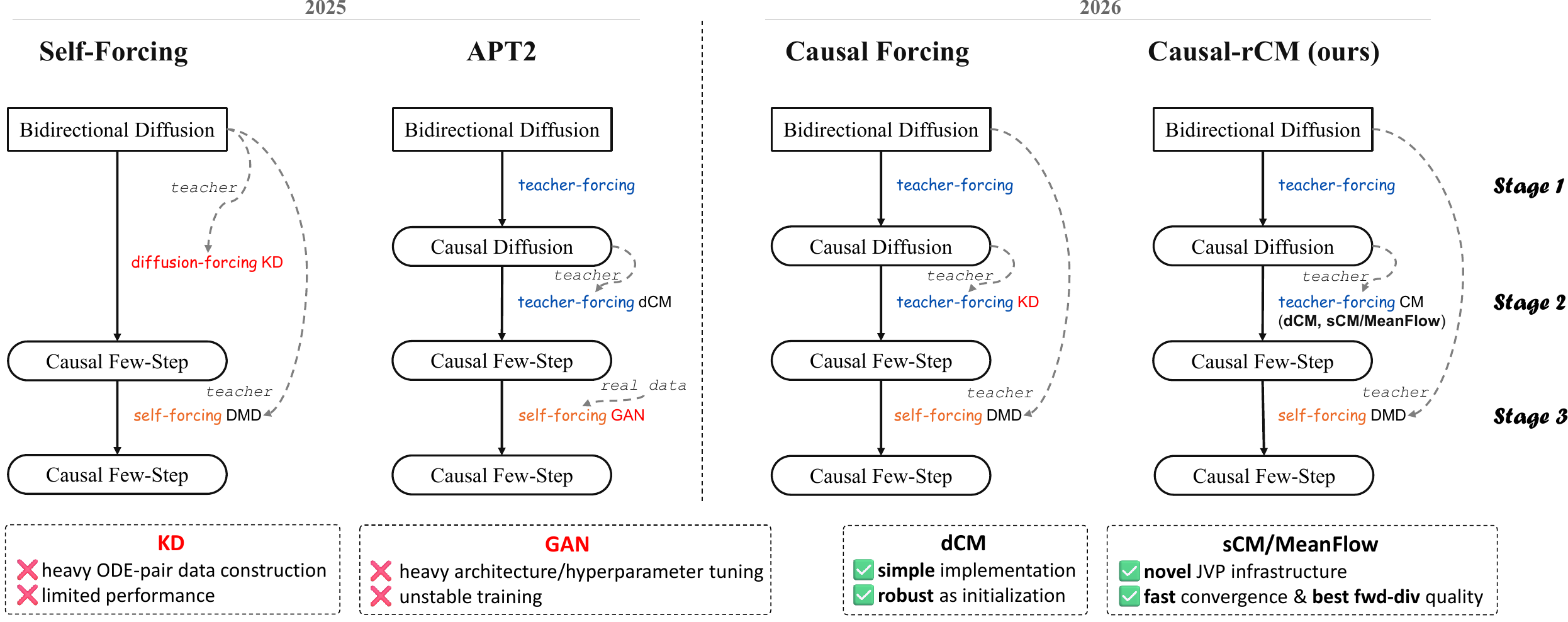}
    \caption{\centering\textbf{Comparison between Causal-rCM and other approaches.}}
    \label{fig:comparison}
\end{figure}

%% file: figures/diagonal.tex
\begin{figure}[htbp]
    \centering
    \includegraphics[width=0.75\textwidth]{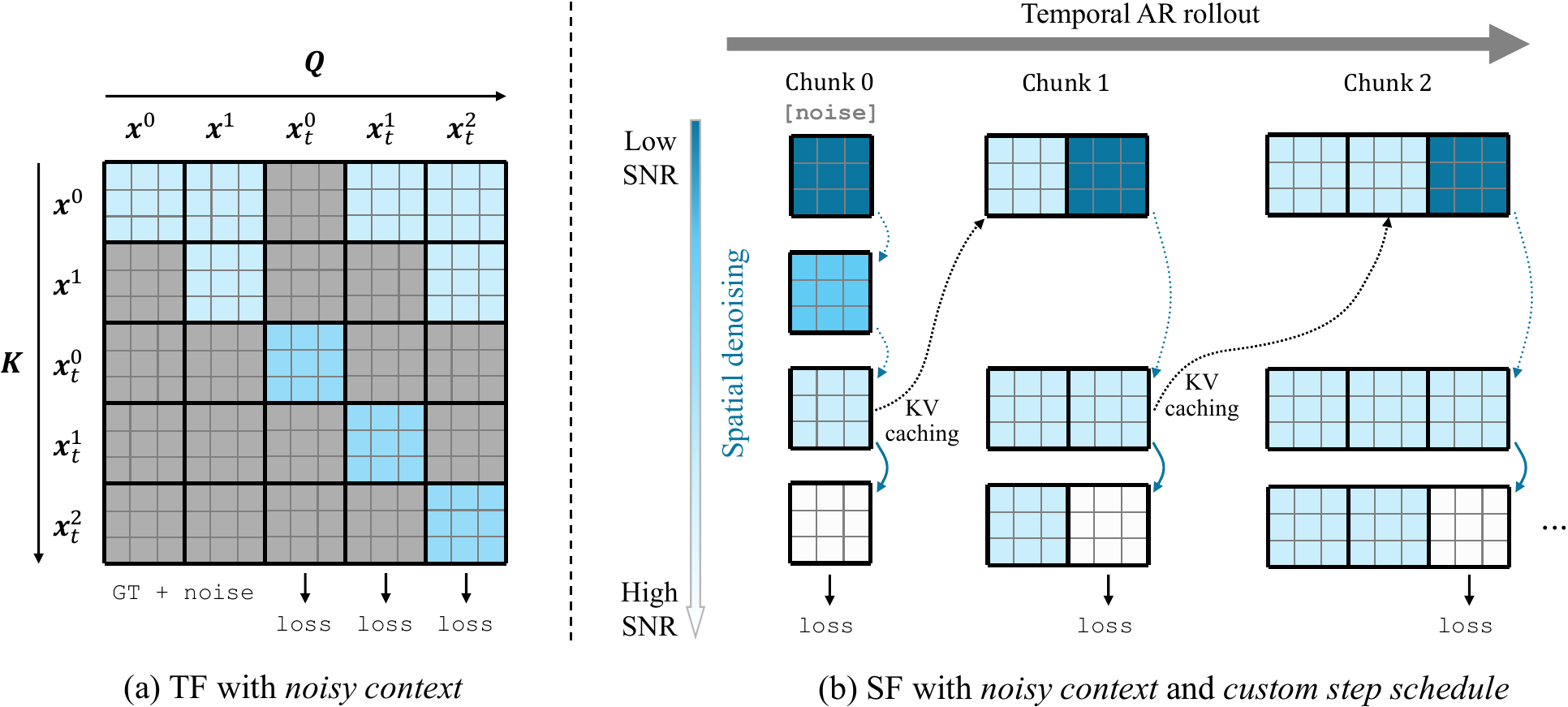}
    \caption{\centering\textbf{Adaptation to acceleration techniques: noisy context and custom step schedule.}}
    \label{fig:diagonal}
\end{figure}

%% file: tables/infra.tex
\providecommand{\infraY}{\ding{51}}
\providecommand{\infraN}{\ding{55}}
\providecommand{\infraP}{\ensuremath{\triangle}}
\providecommand{\infraNA}{--}
\providecommand{\infraA}[2]{#1\textsuperscript{#2}}

\colorlet{llgray}{lightgray!40}

\begin{table*}[ht]
\centering
\scriptsize
\setlength{\tabcolsep}{3.0pt}
\renewcommand{\arraystretch}{1.15}

\caption{\centering \textbf{Implementation-level comparison of autoregressive video diffusion codebases.}}
\label{tab:wan_causal_distillation_infra}

\resizebox{\textwidth}{!}{%
\begin{tabular}{@{}lccccccccccccccc@{}}
\toprule
\multirow{2}{*}{Codebase}
& \multicolumn{2}{c}{Recipe Scope}
& \multicolumn{4}{c}{Algorithmic Recipes}
& \multicolumn{4}{c}{Bidirectional Infra}
& \multicolumn{5}{c}{Causal Infra} \\
\cmidrule(lr){2-3}
\cmidrule(lr){4-7}
\cmidrule(lr){8-11}
\cmidrule(lr){12-16}
&
Bi. & Causal
& TF
& DF
& SF
& Replayed
& FSDP2
& CP/SP
& SAC
& JVP 
& FSDP2
& CP/SP
& SAC 
& JVP
& KV Cache \\
\midrule

Self-Forcing~\citep{huang2025selfforcing}
& \infraN
& \infraY
& \infraY
& \infraY
& \infraY
& \infraN
& \infraN
& \infraN
& \infraN
& \infraN
& \infraA{\infraP}{v1}
& \infraN
& \infraA{\infraP}{AC}
& \infraN
& \infraA{\infraY}{post} \\

FastVideo~\citep{fastvideo_codebase}
& \infraY
& \infraY
& \infraN
& \infraY
& \infraY
& \infraN
& \infraY
& \infraA{\infraY}{F-U}
& \infraY
& \infraN
& \infraY
& \infraN
& \infraY
& \infraN
& \infraA{\infraY}{post} \\

FastGen~\citep{fastgen2026}
& \infraY
& \infraY
& \infraN
& \infraY
& \infraY
& \infraN
& \infraY
& \infraN
& \infraA{\infraP}{AC}
& \infraN
& \infraY
& \infraN
& \infraA{\infraP}{AC}
& \infraN
& \infraA{\infraY}{post} \\

\cellcolor{llgray}\textbf{(Causal-)rCM}
& \cellcolor{llgray}\infraY
& \cellcolor{llgray}\infraY
& \cellcolor{llgray}\infraY
& \cellcolor{llgray}\infraY
& \cellcolor{llgray}\infraY
& \cellcolor{llgray}\infraY
& \cellcolor{llgray}\infraY
& \cellcolor{llgray}\infraA{\infraY}{F-U}
& \cellcolor{llgray}\infraY
& \cellcolor{llgray}\infraY
& \cellcolor{llgray}\infraY
& \cellcolor{llgray}\infraA{\infraY}{F-U}
& \cellcolor{llgray}\infraY
& \cellcolor{llgray}\infraY
& \cellcolor{llgray}\infraA{\infraY}{pre/post} \\

\bottomrule
\end{tabular}%
}

\vspace{0.35em}
\begin{minipage}{0.98\textwidth}
\footnotesize
\textit{Notes.}
\infraY{}: supported;
\infraN{}: not found;
\infraP{}: partial, unclear, or path-dependent.

\textbf{TF}: teacher-forcing implemented as \texttt{[clean frames, noisy frames]} concatenation with a special causal mask.

\textbf{DF}: diffusion-forcing with ordinary block-causal masking.

\textbf{SF}: self-forcing with self-rollout / KV-cache-style training execution.

\textbf{Replayed}: replayed back-propagation technique that avoids storing the entire computation
graph during self-rollout.

\textbf{FSDP2}: Fully Sharded Data Parallel v2. \textsuperscript{v1}: FSDP1-only support.

\textbf{CP/SP}: context/sequence parallel. \textsuperscript{T}: temporal/frame axis;
\textsuperscript{F}: flattened video-token axis, e.g., flattened \texttt{T$\times$H$\times$W} patch tokens.
\textsuperscript{U}: DeepSpeed-Ulysses;
\textsuperscript{R}: Ring Attention;
\textsuperscript{UR}: Ulysses--Ring hybrid (USP).

\textbf{SAC}: selective activation checkpointing. \textsuperscript{AC}: activation checkpointing, but not clearly op-level SAC.

\textbf{JVP}: Jacobian-vector-product. Base operator for continuous-time consistency model (sCM/MeanFlow).

\textbf{KV Cache}: causal self-attention KV cache. \textsuperscript{pre}: K is cached before RoPE; 
\textsuperscript{post}: K is cached after RoPE.
\end{minipage}

\end{table*}

%% file: tables-exp/exp-detail.tex
\begin{table}[ht]
\centering
\caption{\centering\label{tab:training-config}\textbf{Training configurations for Causal-rCM on Wan2.1 T2V.}}
\resizebox{\linewidth}{!}{
\begin{tabular}{lccccc}
\toprule
\textbf{Configuration}
& \multicolumn{2}{c}{\textbf{Stage 1}}
& \multicolumn{2}{c}{\textbf{Stage 2}}
& \multicolumn{1}{c}{\textbf{Stage 3}} \\
\cmidrule(lr){2-3} \cmidrule(lr){4-5} \cmidrule(lr){6-6}
& \makecell{\textbf{Wan2.1-1.3B}\\\textbf{TF/DF}}
& \makecell{\textbf{Wan2.1-14B}\\\textbf{TF/DF}}
& \makecell{\textbf{Wan2.1-1.3B}\\\textbf{TF-dCM}}
& \makecell{\textbf{Wan2.1-1.3B}\\\textbf{TF-sCM}}
& \makecell{\textbf{Wan2.1-1.3B}\\\textbf{SF-DMD}} \\
\midrule
Global batch size & 256 & 64 & 32 & 32 & 64 \\
Context parallel size & 1 & 8 & 4 & 4 & 4 \\
Student optimizer & \makecell{AdamW\\lr $=10^{-5}$\\$\beta=(0.9,0.999)$\\wd $=0.01$} & \makecell{AdamW\\lr $=10^{-5}$\\$\beta=(0.9,0.999)$\\wd $=0.01$} & \makecell{AdamW\\lr $=2\!\times\!10^{-6}$\\$\beta=(0,0.999)$\\wd $=0.01$} & \makecell{AdamW\\lr $=2\!\times\!10^{-6}$\\$\beta=(0,0.999)$\\wd $=0.01$} & \makecell{AdamW\\lr $=2\!\times\!10^{-6}$\\$\beta=(0,0.999)$\\wd $=0.01$} \\
Fake-score optimizer & -- & -- & -- & -- & \makecell{AdamW\\lr $=4\!\times\!10^{-7}$\\$\beta=(0,0.999)$\\wd $=0.01$} \\
CFG scale & -- & -- & 3.0 & 3.0 & 5.0 \\
Time sampling / weighting & \makecell{TF: $p_G=\mathrm{UniformShift}(5)$,\\shared $t$, Gaussian-bell weight;\\DF: $p_G=\mathrm{UniformShift}(5)$,\\random per-chunk $t$, no weight} & \makecell{TF: $p_G=\mathrm{UniformShift}(5)$,\\shared $t$, Gaussian-bell weight;\\DF: $p_G=\mathrm{UniformShift}(5)$,\\random per-chunk $t$, no weight} & \makecell{uniform RF grid with\\shift $=3$, steps $=48$, skip $=1$} & \makecell{$p_G=\mathrm{LogitNormal}$\\$(\mu=-0.8,\sigma=1.6)$} & \makecell{$p_D=\mathrm{UniformShift}(5)$} \\
Specific hyperparameters & -- & -- & -- & tangent warmup $=1000$ & \makecell{max rollout steps $=4$\\student update freq. $=6$} \\
Training iterations & 30k & 30k & 10k & 1k & varies \\
\bottomrule
\end{tabular}
}
\end{table}

%% file: tables-exp/main.tex
\colorlet{llgray}{lightgray!40}

\begin{table}[ht]
\centering
\small
\setlength{\tabcolsep}{3pt}
\caption{\centering\label{tab:main_results}\textbf{Main streaming video generation results on Wan2.1 T2V.}}
\resizebox{\linewidth}{!}{
\begin{tabular}{lcccccccc}
\toprule
\textbf{Method} & \textbf{NFE} & \textbf{Total Score$\uparrow$} & \textbf{Quality Score$\uparrow$} & \textbf{Semantic Score$\uparrow$} & \textbf{Throughput$\uparrow$} & \textbf{First Latency$\downarrow$} & \textbf{Second Latency$\downarrow$} & \textbf{SF-DMD iters} \\
 & & & & & \textbf{(FPS)} & \textbf{(s)} & \textbf{(s)} & \\
\midrule
\rowcolor{llgray}
\multicolumn{9}{l}{\textit{Bidirectional}} \\
Wan2.1-1.3B & 50$\times$2 & 82.78 & 83.44 & 80.13 & 0.72 & -- & -- & -- \\
Wan2.1-14B & 50$\times$2 & 83.35 & 83.97 & 80.88 & 0.18 & -- & -- & -- \\
\midrule
\rowcolor{llgray}
\multicolumn{9}{l}{\textit{Frame-wise (\textit{c1-1})}} \\
Causal Forcing (4-step) & 5 & 81.56 & 82.59 & 77.44 & 8.3 & 0.40 & 0.46 & -- \\
\textbf{Causal-rCM (4-step)} & 5 & \underline{84.29} & 85.27 & 80.36 & 8.3 & 0.40 & 0.46 & 1200 \\
\textbf{Causal-rCM (2-step)} & 3 & \textbf{84.63} & 85.46 & 81.31 & 12.2 & 0.40 & 0.31 & 3000 \\
\textbf{Causal-rCM (2-step, noisy ctx)} & 2 & 83.11 & 83.55 & 81.37 & 15.9 & 0.40 & 0.23 & 1500 \\
\textbf{Causal-rCM (1-step)} & 2 & \textbf{84.63} & 85.54 & 81.01 & 15.9 & 0.40 & 0.23 & 3000 \\
\midrule
\rowcolor{llgray}
\multicolumn{9}{l}{\textit{Chunk-wise (\textit{c3-3})}} \\
Self-Forcing (4-step) & 5 & 83.76 & 84.53 & 80.68 & 17.4 & 0.57 & 0.64 & -- \\
LongLive (4-step) & 5 & 83.62 & 84.36 & 80.69 & 17.4 & 0.57 & 0.64 & -- \\
Causal Forcing (4-step) & 5 & 83.96 & 84.94 & 80.04 & 17.4 & 0.57 & 0.64 & -- \\
AnyFlow (4-step) & 5 & 84.31 & 85.15 & 80.94 & 17.4 & 0.57 & 0.64 & -- \\
\textbf{Causal-rCM (4-step)} & 5 & \textbf{84.37} & 85.02 & 81.73 & 17.4 & 0.57 & 0.64 & 1250 \\
\textbf{Causal-rCM (2-step)} & 3 & \underline{84.30} & 85.04 & 81.36 & 22.2 & 0.57 & 0.49 & 2500 \\
\textbf{Causal-rCM (2-step, noisy ctx)} & 2 & 84.24 & 84.96 & 81.36 & 25.6 & 0.57 & 0.41 & 1750 \\
\textbf{Causal-rCM (1-step)} & 2 & 84.01 & 84.71 & 81.22 & 25.6 & 0.57 & 0.41 & 3000 \\
\bottomrule
\end{tabular}
}
\end{table}

%% file: figures/dcm_vs_scm.tex
\begin{figure}[htbp]
    \centering
    \includegraphics[width=0.5\textwidth]{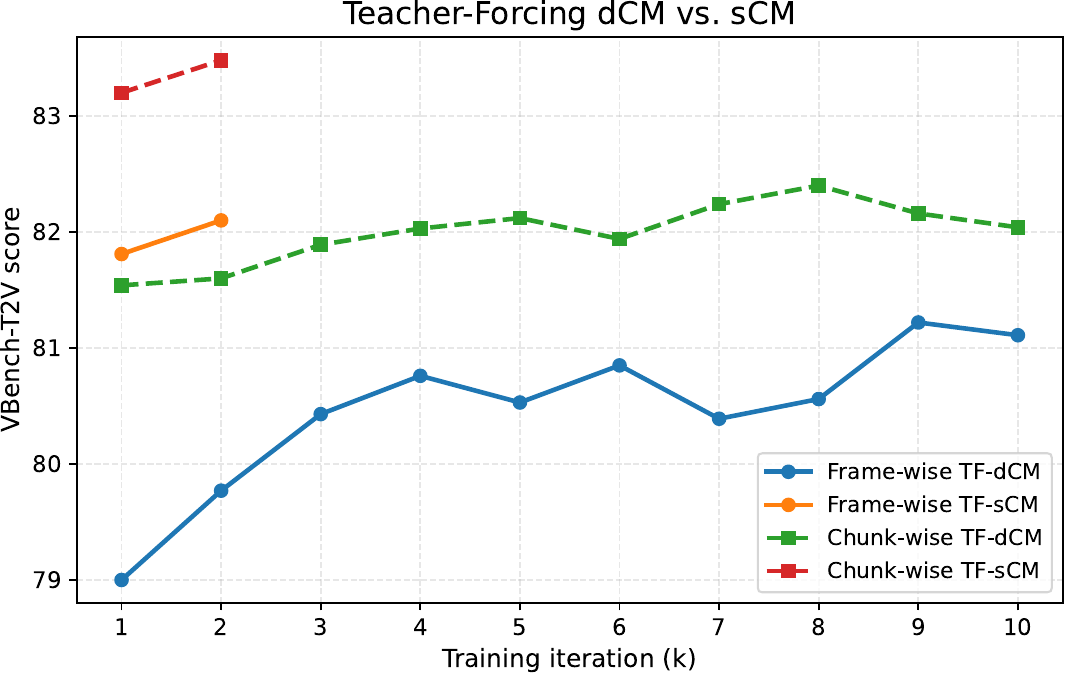}
    \caption{\centering\textbf{Training curves of TF-dCM and TF-sCM.}}
    \label{fig:dcm-vs-scm}
\end{figure}

%% file: tables-exp/ablation.tex
\colorlet{llgray}{lightgray!40}

\begin{table}[t]
\centering
\small
\setlength{\tabcolsep}{6pt}
\caption{\centering\label{tab:ablation}\textbf{Ablation of initialization strategies for 4-step SF-DMD.}}
\resizebox{0.65\linewidth}{!}{
\begin{tabular}{lcccc}
\toprule
\textbf{Initialization} & \textbf{Total Score$\uparrow$} & \textbf{Quality Score$\uparrow$} & \textbf{Semantic Score$\uparrow$} & \textbf{SF-DMD iterations} \\
\midrule
\rowcolor{llgray}
\multicolumn{5}{l}{\textit{Frame-wise (\textit{c1-1})}} \\
DF & 83.11 & 83.85 & 80.16 & 2000 \\
TF & 82.62 & 83.62 & 78.61 & 1000 \\
DF-KD & 80.59 & 80.41 & 81.32 & 2000 \\
TF-KD & 83.49 & 84.50 & 79.43 & 1250 \\
TF-dCM & \textbf{84.29} & 85.27 & 80.36 & 1200 \\
TF-sCM & \underline{83.84} & 84.67 & 80.55 & 1000 \\
\midrule
\rowcolor{llgray}
\multicolumn{5}{l}{\textit{Chunk-wise (\textit{c3-3})}} \\
DF & \textbf{84.80} & 85.58 & 81.65 & 1500 \\
TF & \textbf{84.95} & 85.82 & 81.47 & 1000 \\
DF-KD & 83.61 & 84.10 & 81.68 & 1500 \\
TF-KD & 83.79 & 84.41 & 81.30 & 1000 \\
TF-dCM & \underline{84.33} & 85.22 & 80.75 & 3200 \\
TF-sCM & \underline{84.37} & 85.02 & 81.73 & 1250 \\
\bottomrule
\end{tabular}
}
\end{table}

%% file: figures/vbench_curve.tex
\begin{figure}[t]
\centering
\begin{subfigure}[t]{0.49\textwidth}
\centering
\includegraphics[width=\linewidth]{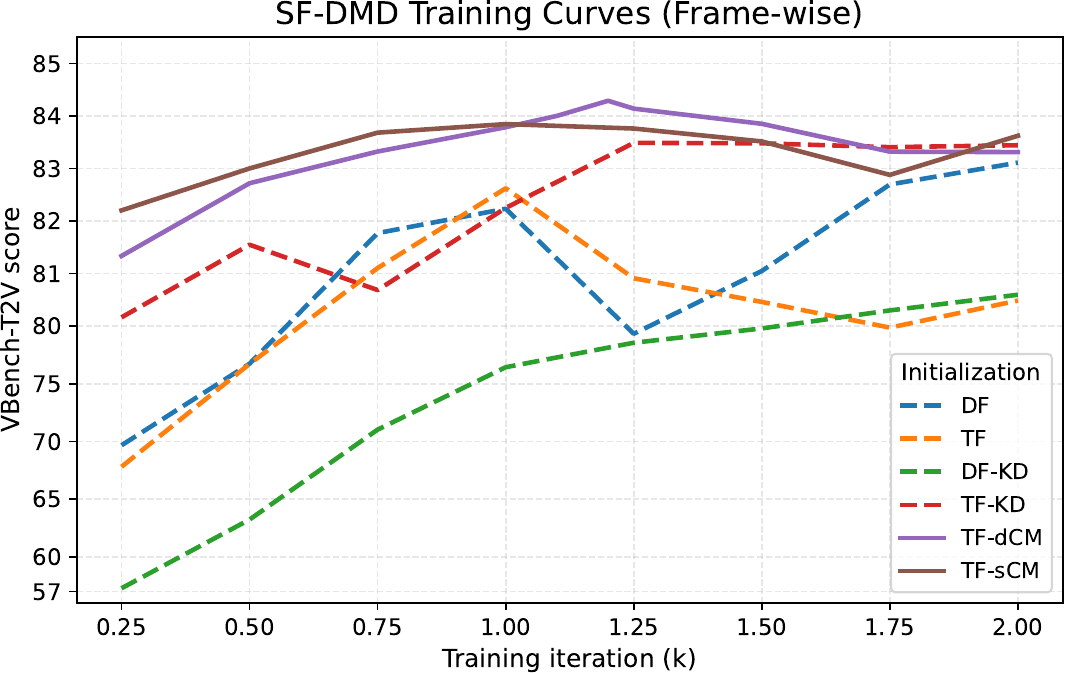}
\caption{\centering Frame-wise}
\label{fig:vbench-curve1}
\end{subfigure}
\hfill
\begin{subfigure}[t]{0.49\textwidth}
\centering
\includegraphics[width=\linewidth]{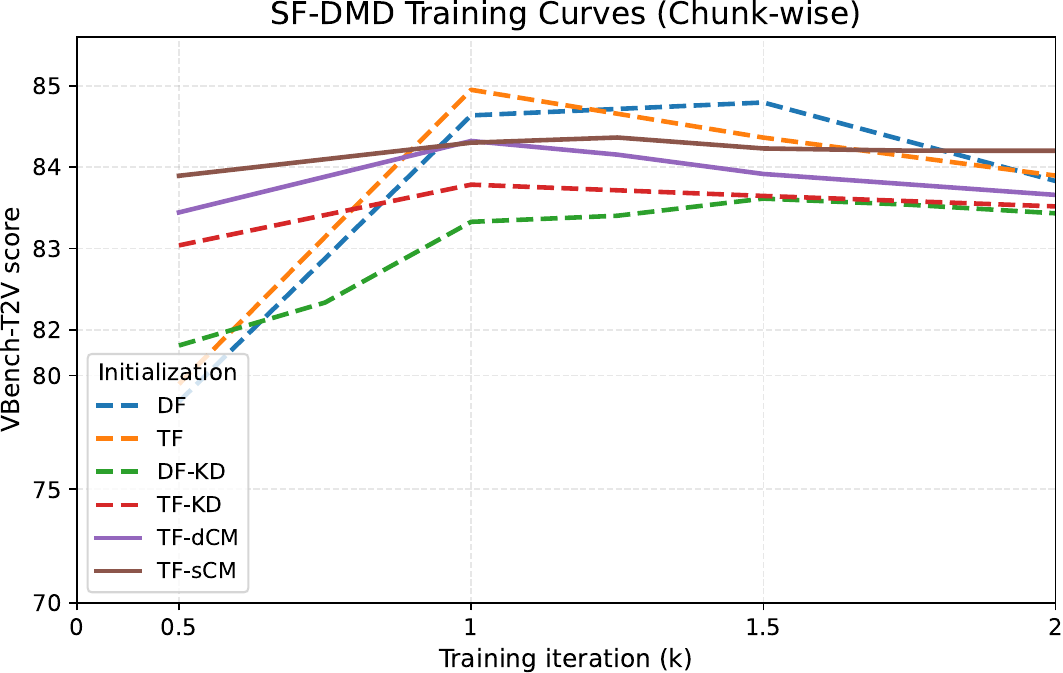}
\caption{\centering Chunk-wise}
\label{fig:vbench-curve2}
\end{subfigure}
\vspace{-.2cm}
\caption{\centering\textbf{SF-DMD training curves with different initialization strategies.}}
\label{fig:vbench-curve}
\vspace{-.4cm}
\end{figure}

%% file: figures/oversmooth.tex
\begin{figure}[t]
\centering
\begin{subfigure}[t]{0.24\textwidth}
\centering
\includegraphics[width=\linewidth]{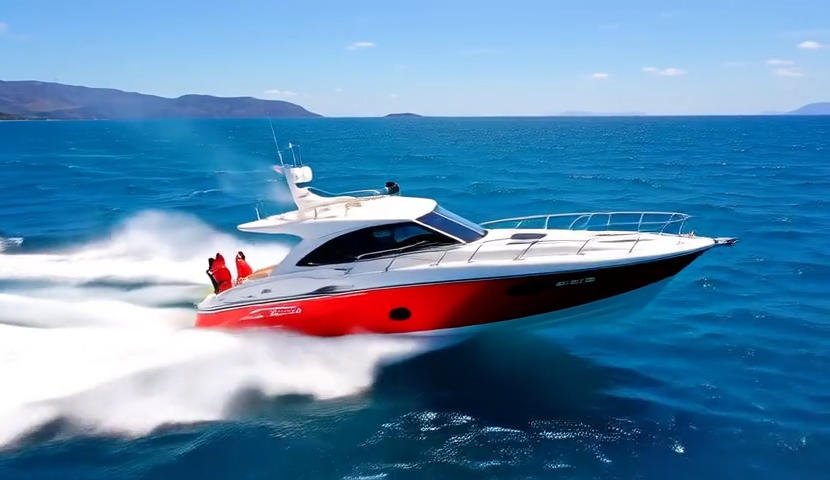}
\caption{\centering DF + SF-DMD}
\end{subfigure}
\hfill
\begin{subfigure}[t]{0.24\textwidth}
\centering
\includegraphics[width=\linewidth]{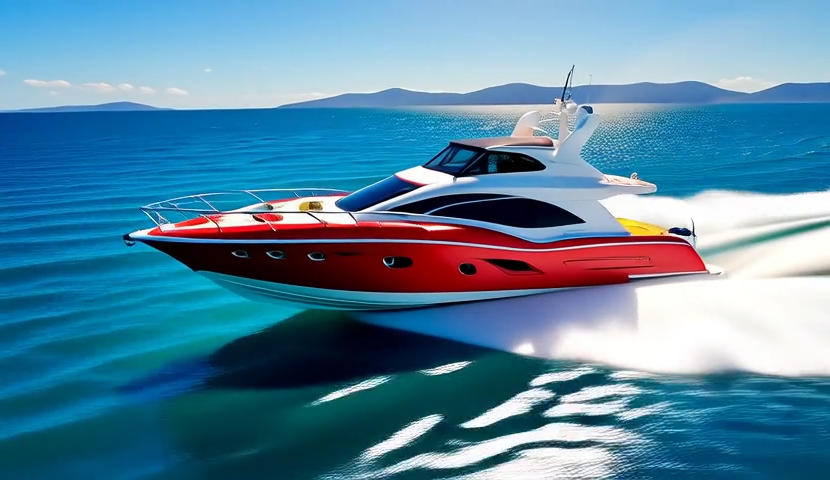}
\caption{\centering TF + SF-DMD}
\end{subfigure}
\hfill
\begin{subfigure}[t]{0.24\textwidth}
\centering
\includegraphics[width=\linewidth]{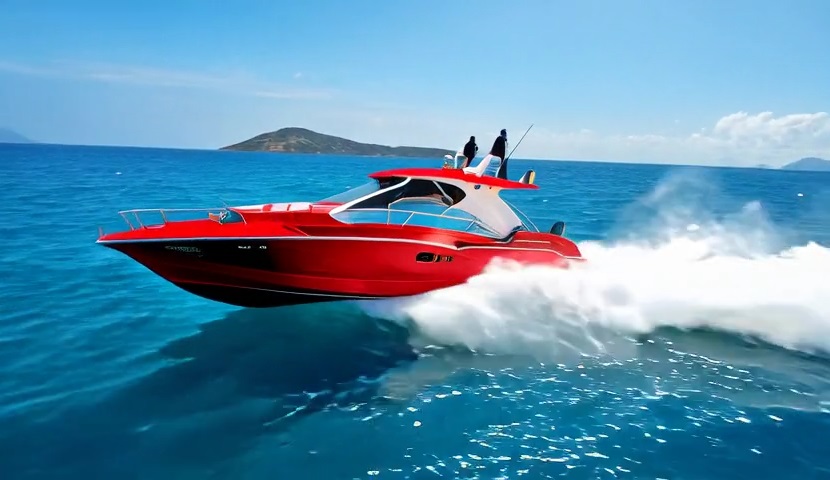}
\caption{\centering TF-dCM + SF-DMD}
\end{subfigure}
\hfill
\begin{subfigure}[t]{0.24\textwidth}
\centering
\includegraphics[width=\linewidth]{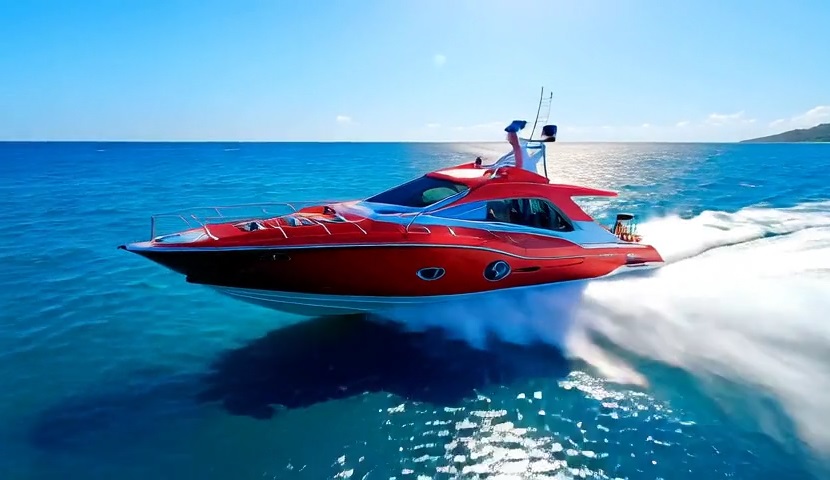}
\caption{\centering TF-sCM + SF-DMD}
\end{subfigure}
\vspace{-.2cm}
\caption{\textbf{Visualizations of chunk-wise SF-DMD under different initialization strategies.} DF/TF initialization leads to higher VBench-T2V scores while suffering from overly smooth textures and lacking fine-grained details.}
\label{fig:oversmooth}
\end{figure}

%% file: figures/interactive_overview.tex
\begin{figure}[htbp]
    \centering
    \includegraphics[width=0.7\textwidth]{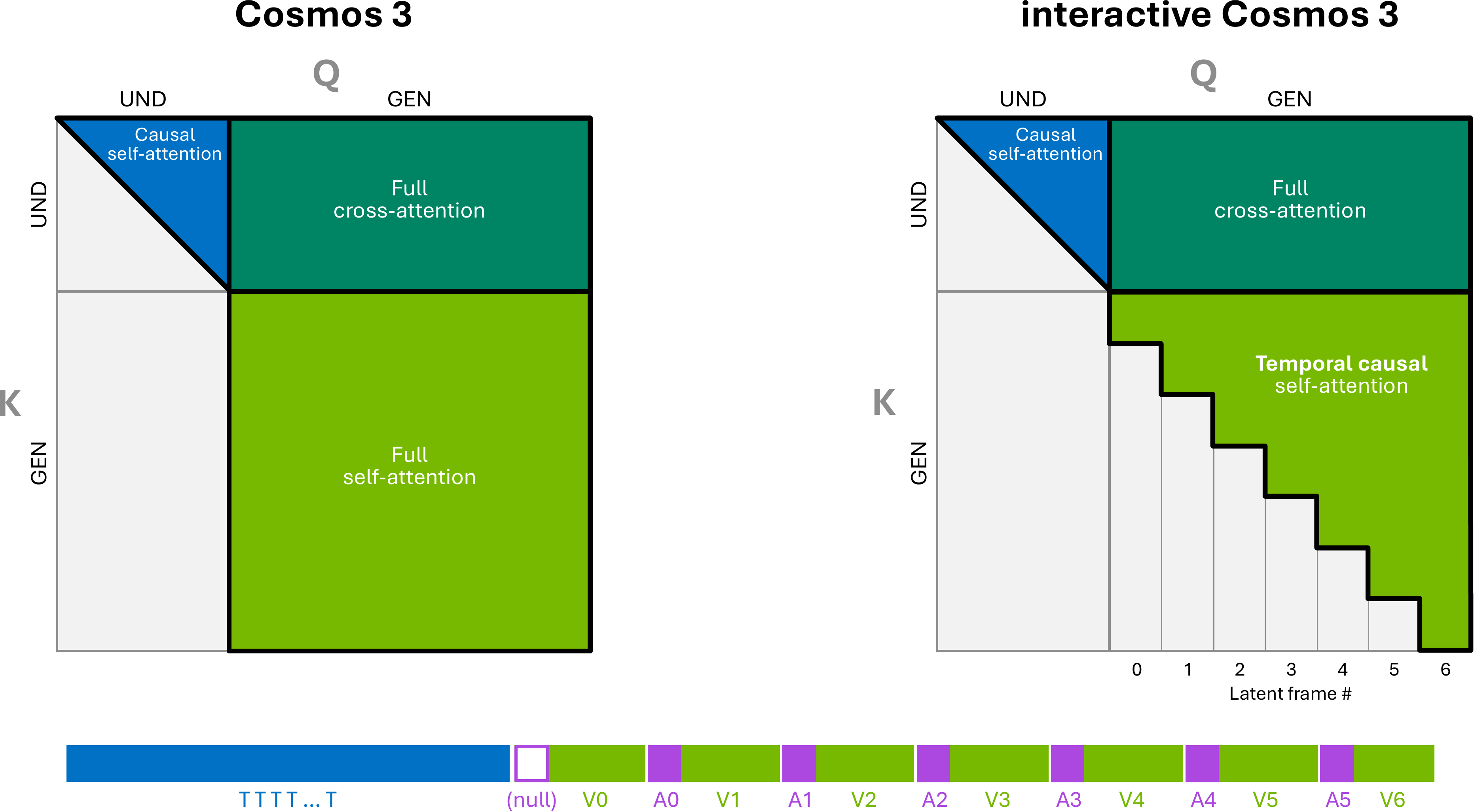}
    \caption{\textbf{From Cosmos 3 to interactive Cosmos 3.} Cosmos 3 uses causal self-attention for UND tokens, full cross-attention from GEN to UND tokens, and bidirectional self-attention within GEN tokens. Interactive Cosmos 3 preserves the UND-GEN attention structure but replaces GEN self-attention with temporal-causal attention over latent-frame supertokens. In the forward-dynamics layout, \(V_i\) denotes a vision supertoken, \(A_i\) controls the transition to \(V_{i+1}\), and a null action token is inserted before \(V_0\) to keep a uniform token layout.}
    \label{fig:interactive-overview}
\end{figure}

%% file: figures/interactive_av.tex
\begin{figure}[htbp]
    \centering
    \includegraphics[width=1.0\textwidth]{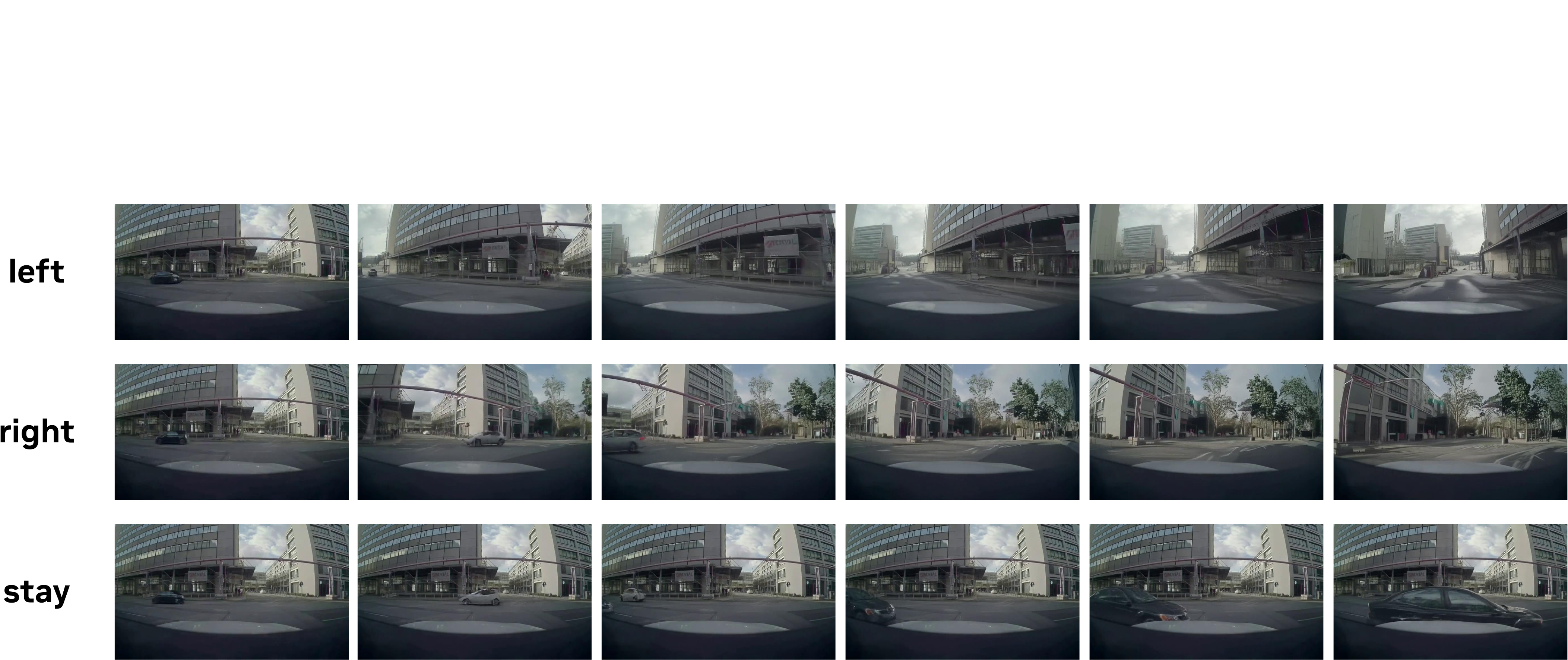}
    \caption{\centering\textbf{Cosmos 3 interactive generation on autonomous-driving scenarios conditioned on the action of the vehicle ego-motion.}}
    \label{fig:interactive-av}
\end{figure}

%% file: tables/distillation_pipeline.tex
\begin{table*}[t]
\centering
\small
\setlength{\tabcolsep}{4pt}
\renewcommand{\arraystretch}{1.15}
\caption{\centering\label{tab:distillation_pipeline_insight}\textbf{A high-level view of CM/CTM distillation recipes.}}
\resizebox{\linewidth}{!}{
\begin{tabular}{lllp{7.2cm}}
\toprule
\textbf{Route} & \textbf{Setting} & \textbf{Ultimate pipeline} & \textbf{Related works} \\
\midrule
CM & Bidirectional &
dCM $\rightarrow$ sCM $\rightarrow$ DMD/GAN $(+\,$CM/on-policy CM$)$ &
APT~\citep{lin2025diffusion}: dCM $\rightarrow$ GAN;

rCM~\citep{zheng2025large}: sCM $+$ DMD. \\
CM & Causal &
TF-dCM $\rightarrow$ TF-sCM $\rightarrow$ SF-DMD/SF-GAN $(+\,$TF-CM/SF-CM$)$ &
APT2~\citep{lin2025autoregressive}: TF-dCM $\rightarrow$ SF-GAN;

CF++~\citep{zhao2026causal}: TF-dCM $\rightarrow$ SF-DMD;

Causal-rCM (ours): TF-dCM/TF-sCM $\rightarrow$ SF-DMD.\\
CTM & Bidirectional &
MeanFlow (FD) $\rightarrow$ MeanFlow (JVP) $\rightarrow$ DMD/GAN $(+\,$MeanFlow/on-policy MeanFlow$)$ &
Transition Matching~\citep{nie2026transition}: MeanFlow (FD) $\rightarrow$ DMD2-v with flow-head rollout;

AnyFlow~\citep{gu2026anyflow}: MeanFlow (FD) $\rightarrow$ DMD + on-policy MeanFlow (FD).\\
CTM & Causal &
TF-MeanFlow (FD) $\rightarrow$ TF-MeanFlow (JVP) $\rightarrow$ SF-DMD/SF-GAN $(+\,$TF-MeanFlow/SF-MeanFlow$)$ &
AnyFlow~\citep{gu2026anyflow}: TF-MeanFlow (FD) $\rightarrow$ SF-DMD + SF-MeanFlow (FD). \\
\bottomrule
\end{tabular}
}
\vspace{2pt}
\caption*{\footnotesize \textit{Notes.}
MeanFlow (FD) denotes finite-difference-estimated MeanFlow, and MeanFlow (JVP) denotes the exact JVP-based MeanFlow.

$\rightarrow$: different stages; $+$: joint training.}
\end{table*}

%% file: sections/appendix_trigflow_scm_vs_rf_scm.tex
\section{Theoretical Analysis of TrigFlow-sCM and RF-sCM}
\label{app:trigflow-scm-vs-rf-scm}

This section compares two implementations of continuous-time consistency distillation for an RF-native velocity predictor: (i) applying a TrigFlow wrapper to the RF velocity predictor and then using the TrigFlow-sCM objective~\citep{zheng2025large}, and (ii) directly writing the sCM objective under the RF schedule. Despite that different diffusion noise schedules (e.g., TrigFlow and RF) are equivalent and mutually convertible~\citep{zheng2023improved} up to a scaling factor, we show that they result in generally different normalized MSE training objectives for sCM. The difference comes from the input-output scaling of the TrigFlow wrapper, the tangent normalization, and the finite-precision evaluation order of JVPs.

\paragraph{RF and TrigFlow coordinates.}
Let \(u\in[0,1]\) denote the RF time and let \(\tau\in[0,\pi/2]\) denote the TrigFlow time. Define
\[
    C=\cos\tau,\qquad S=\sin\tau,\qquad
    Z=C+S,\qquad b=Z^{-1},\qquad
    u=\frac{S}{Z}.
\]
The RF and TrigFlow forward processes are
\[
    \xv_u=(1-u)\xv_0+u\epsilonv,
    \qquad
    \xv_\tau=C\xv_0+S\epsilonv.
\]
They are related by a time-dependent state scaling:
\begin{equation}
\label{eq:app-trig-rf-path-relation}
    \xv_\tau=Z\xv_u,
    \qquad
    \xv_u=b\xv_\tau.
\end{equation}
Let \(\vv_\theta(\xv_u,u)\) be an RF velocity predictor and let \(\mathbf V(\xv_u,u)=\vv_{\mathrm{teacher}}(\xv_u,u)\) denote the RF teacher velocity. The direct RF consistency map is
\begin{equation}
\label{eq:app-rf-consistency-map}
    \fv_\theta^{\mathrm{RF}}(\xv_u,u)
    =
    \xv_u-u\vv_\theta(\xv_u,u).
\end{equation}

\paragraph{TrigFlow wrapper as input-output transforms.}
The TrigFlow wrapper around the RF velocity predictor can be written as
\begin{equation}
\label{eq:app-trig-wrapper}
    \Fv_\theta^{\mathrm{trig}}(\xv_\tau,\tau)
    =
    (C-S)\xv_u+b\,\vv_\theta(\xv_u,u),
    \qquad
    \xv_u=b\xv_\tau,\quad u=\frac{S}{Z}.
\end{equation}
Equivalently, the wrapper first applies the input transform \((\xv_\tau,\tau)\mapsto(\xv_u,u)\), evaluates the RF velocity predictor, and then applies the output transform
\[
    \vv_\theta(\xv_u,u)
    \mapsto
    (C-S)\xv_u+b\,\vv_\theta(\xv_u,u).
\]
Under the TrigFlow preconditioning
\begin{equation}
\label{eq:app-trig-consistency-map}
    \fv_\theta^{\mathrm{trig}}(\xv_\tau,\tau)
    =
    C\xv_\tau-S\Fv_\theta^{\mathrm{trig}}(\xv_\tau,\tau),
\end{equation}
substituting Eqn.~\ref{eq:app-trig-wrapper} gives
\begin{align}
    \fv_\theta^{\mathrm{trig}}(\xv_\tau,\tau)
    &=
    CZ\xv_u
    -
    S\left[(C-S)\xv_u+b\vv_\theta(\xv_u,u)\right] \nonumber\\
    &=
    (C^2+CS-CS+S^2)\xv_u
    -
    \frac{S}{Z}\vv_\theta(\xv_u,u) \nonumber\\
    &=
    \xv_u-u\vv_\theta(\xv_u,u)
    =
    \fv_\theta^{\mathrm{RF}}(\xv_u,u).
\label{eq:app-trig-rf-same-map}
\end{align}
Therefore, the TrigFlow wrapper and the direct RF parameterization define the same consistency map after the change of variables in Eqn.~\ref{eq:app-trig-rf-path-relation}.

\paragraph{Direct RF-sCM tangent.}
The RF teacher ODE is
\[
    \frac{\dm \xv_u}{\dm u}=\mathbf V(\xv_u,u).
\]
For the stop-gradient network \(\theta^-\), define the RF JVP
\begin{equation}
\label{eq:app-rf-jvp}
    \mathbf J_{\theta^-}^{\mathrm{RF}}
    =
    \texttt{JVP}
    \left(
    \vv_{\theta^-};
    (\xv_u,u),
    (\mathbf V,1)
    \right)
    =
    \nabla_{\xv_u}\vv_{\theta^-}\,\mathbf V
    +
    \partial_u\vv_{\theta^-}.
\end{equation}
The tangent of the RF consistency map is
\begin{align}
    \mathbf h_{\mathrm{RF}}
    &=
    \frac{\dm}{\dm u}
    \left[
    \xv_u-u\vv_{\theta^-}(\xv_u,u)
    \right] \nonumber\\
    &=
    \mathbf V-\vv_{\theta^-}-u\,\mathbf J_{\theta^-}^{\mathrm{RF}} .
\label{eq:app-rf-tangent}
\end{align}
The direct RF-sCM objective can thus be written as
\begin{equation}
\label{eq:app-rf-scm}
    \Lc_{\mathrm{RF\text{-}sCM}}
    =
    \E
    \left[
    \left\|
    \Delta\vv
    -
    \frac{
    w_{\mathrm{RF}}(u)\mathbf h_{\mathrm{RF}}
    }{
    w_{\mathrm{RF}}^2(u)\|\mathbf h_{\mathrm{RF}}\|_2^2+c
    }
    \right\|_2^2
    \right],
    \qquad
    \Delta\vv=\vv_\theta-\vv_{\theta^-}.
\end{equation}
\(\Delta\vv\) is zero in the forward value, but it still indicates the output coordinate with respect to which gradients are taken.

\paragraph{TrigFlow JVP through the input and output transforms.}
We next rewrite the TrigFlow-sCM JVP in the RF coordinates. Along the TrigFlow teacher trajectory,
\begin{equation}
\label{eq:app-time-jacobian}
    \frac{\dm u}{\dm\tau}=b^2.
\end{equation}
Using \(\xv_u=b\xv_\tau\) and \(\frac{\dm\xv_\tau}{\dm\tau}=\Fv_{\mathrm{teacher}}^{\mathrm{trig}}(\xv_\tau,\tau)\), where
\[
    \Fv_{\mathrm{teacher}}^{\mathrm{trig}}(\xv_\tau,\tau)
    =
    (C-S)\xv_u+b\mathbf V,
\]
we obtain
\begin{align}
    \frac{\dm\xv_u}{\dm\tau}
    &=
    \frac{\dm}{\dm\tau}(b\xv_\tau)
    =
    \dot b\,\xv_\tau+b\frac{\dm\xv_\tau}{\dm\tau} \nonumber\\
    &=
    \dot b\,Z\xv_u+b\left[(C-S)\xv_u+b\mathbf V\right].
\end{align}
Since
\[
    \dot b
    =
    \frac{\dm}{\dm\tau}\frac{1}{C+S}
    =
    -\frac{C-S}{Z^2}
    =
    -(C-S)b^2,
\]
the explicit state terms cancel and
\begin{equation}
\label{eq:app-input-transform-tangent}
    \frac{\dm\xv_u}{\dm\tau}=b^2\mathbf V.
\end{equation}
Thus the JVP direction entering the RF velocity predictor inside the TrigFlow wrapper is
\begin{equation}
\label{eq:app-scaled-rf-jvp-direction}
    \left(
    \frac{\dm\xv_u}{\dm\tau},
    \frac{\dm u}{\dm\tau}
    \right)
    =
    b^2(\mathbf V,1).
\end{equation}
In exact arithmetic,
\begin{equation}
\label{eq:app-scaled-rf-jvp}
    \texttt{JVP}
    \left(
    \vv_{\theta^-};
    (\xv_u,u),
    b^2(\mathbf V,1)
    \right)
    =
    b^2\mathbf J_{\theta^-}^{\mathrm{RF}}.
\end{equation}

The output transform in Eqn.~\ref{eq:app-trig-wrapper} contains explicit \(\tau\)-dependent coefficients. Therefore the JVP of the wrapped velocity is
\begin{align}
    \frac{\dm}{\dm\tau}
    \Fv_{\theta^-}^{\mathrm{trig}}
    &=
    \frac{\dm}{\dm\tau}
    \left[
    (C-S)\xv_u+b\vv_{\theta^-}(\xv_u,u)
    \right] \nonumber\\
    &=
    -Z\xv_u
    +(C-S)\frac{\dm\xv_u}{\dm\tau}
    +\dot b\,\vv_{\theta^-}
    +b\,
    \texttt{JVP}
    \left(
    \vv_{\theta^-};
    (\xv_u,u),
    \left(\frac{\dm\xv_u}{\dm\tau},\frac{\dm u}{\dm\tau}\right)
    \right) \nonumber\\
    &=
    -Z\xv_u
    +(C-S)b^2(\mathbf V-\vv_{\theta^-})
    +b^3\mathbf J_{\theta^-}^{\mathrm{RF}} .
\label{eq:app-trig-velocity-jvp}
\end{align}
Now differentiate the TrigFlow consistency map in Eqn.~\ref{eq:app-trig-consistency-map}:
\begin{align}
    \mathbf h_{\mathrm{trig}}
    &=
    \frac{\dm}{\dm\tau}
    \left[
    C\xv_\tau
    -
    S\Fv_{\theta^-}^{\mathrm{trig}}(\xv_\tau,\tau)
    \right] \nonumber\\
    &=
    -S\xv_\tau
    +C\frac{\dm\xv_\tau}{\dm\tau}
    -C\Fv_{\theta^-}^{\mathrm{trig}}
    -S\frac{\dm}{\dm\tau}\Fv_{\theta^-}^{\mathrm{trig}} .
\end{align}
Substituting \(\xv_\tau=Z\xv_u\),
\[
    \frac{\dm\xv_\tau}{\dm\tau}
    =
    (C-S)\xv_u+b\mathbf V,
    \qquad
    \Fv_{\theta^-}^{\mathrm{trig}}
    =
    (C-S)\xv_u+b\vv_{\theta^-},
\]
and Eqn.~\ref{eq:app-trig-velocity-jvp}, we get
\begin{align}
    \mathbf h_{\mathrm{trig}}
    &=
    -SZ\xv_u
    +Cb(\mathbf V-\vv_{\theta^-})
    -S\left[
    -Z\xv_u
    +(C-S)b^2(\mathbf V-\vv_{\theta^-})
    +b^3\mathbf J_{\theta^-}^{\mathrm{RF}}
    \right] \nonumber\\
    &=
    \left[Cb-S(C-S)b^2\right](\mathbf V-\vv_{\theta^-})
    -Sb^3\mathbf J_{\theta^-}^{\mathrm{RF}} .
\end{align}
Using
\[
    Cb-S(C-S)b^2=b^2,
    \qquad
    Sb^3=ub^2,
\]
we obtain the compact relation
\begin{equation}
\label{eq:app-trig-tangent-rf-tangent}
    \mathbf h_{\mathrm{trig}}
    =
    b^2
    \left(
    \mathbf V-\vv_{\theta^-}-u\mathbf J_{\theta^-}^{\mathrm{RF}}
    \right)
    =
    b^2\mathbf h_{\mathrm{RF}}.
\end{equation}
This derivation makes explicit that the TrigFlow wrapper introduces no new RF-JVP structure: after the input and output transforms, the same RF combination
\[
    \mathbf V-\vv_{\theta^-}
    -
    u\,
    \texttt{JVP}
    \left(
    \vv_{\theta^-};
    (\xv_u,u),
    (\mathbf V,1)
    \right)
\]
appears. The only exact-arithmetic difference at the tangent level is the factor \(b^2=Z^{-2}\).

\paragraph{TrigFlow-sCM objective in RF velocity coordinates.}
The TrigFlow-sCM objective is applied in the TrigFlow velocity coordinate. From Eqn.~\ref{eq:app-trig-wrapper},
\begin{equation}
\label{eq:app-output-difference}
    \Delta\Fv^{\mathrm{trig}}
    =
    \Fv_\theta^{\mathrm{trig}}
    -
    \Fv_{\theta^-}^{\mathrm{trig}}
    =
    b(\vv_\theta-\vv_{\theta^-})
    =
    b\Delta\vv.
\end{equation}
Therefore, with
\[
    \mathbf g_{\mathrm{trig}}
    =
    w_{\mathrm{trig}}(\tau)\mathbf h_{\mathrm{trig}}
    =
    w_{\mathrm{trig}}(\tau)b^2\mathbf h_{\mathrm{RF}},
\]
where $w_{\mathrm{trig}}(\tau)$ is taken as $\cos\tau=C$ in sCM~\citep{lu2024simplifying} and rCM~\citep{zheng2025large}, the TrigFlow-sCM loss becomes
\begin{align}
    \Lc_{\mathrm{TrigFlow\text{-}sCM}}
    &=
    \E
    \left[
    \left\|
    \Delta\Fv^{\mathrm{trig}}
    -
    \frac{
    \mathbf g_{\mathrm{trig}}
    }{
    \|\mathbf g_{\mathrm{trig}}\|_2^2+c
    }
    \right\|_2^2
    \right] \nonumber\\
    &=
    \E
    \left[
    \left\|
    b\Delta\vv
    -
    \frac{
    w_{\mathrm{trig}}(\tau)b^2\mathbf h_{\mathrm{RF}}
    }{
    w^2_{\mathrm{trig}}(\tau)b^4\|\mathbf h_{\mathrm{RF}}\|_2^2+c
    }
    \right\|_2^2
    \right] \nonumber\\
    &=
    \E
    \left[
    \frac{1}{Z^2}
    \left\|
    \Delta\vv
    -
    \frac{
    w_{\mathrm{trig}}(\tau)Z^3\mathbf h_{\mathrm{RF}}
    }{
    w^2_{\mathrm{trig}}(\tau)\|\mathbf h_{\mathrm{RF}}\|_2^2+cZ^4
    }
    \right\|_2^2
    \right].
\label{eq:app-trig-scm-rf-coordinate}
\end{align}
By contrast, direct RF-sCM is Eqn.~\ref{eq:app-rf-scm}. Hence the two objectives share the same zero-consistency condition,
\[
    \mathbf h_{\mathrm{trig}}=0
    \quad\Longleftrightarrow\quad
    \mathbf h_{\mathrm{RF}}=0,
\]
but they are not, in general, the same normalized MSE objective.

\paragraph{Effect of tangent normalization.}
The distinction is easiest to see when \(c=0\). Eqn.~\ref{eq:app-rf-scm} gives the RF normalized tangent target
\[
    \mathbf T_{\mathrm{RF}}
    =
    \frac{
    w_{\mathrm{RF}}\mathbf h_{\mathrm{RF}}
    }{
    w_{\mathrm{RF}}^2\|\mathbf h_{\mathrm{RF}}\|_2^2
    }
    =
    \frac{
    \mathbf h_{\mathrm{RF}}
    }{
    w_{\mathrm{RF}}\|\mathbf h_{\mathrm{RF}}\|_2^2
    }.
\]
Eqn.~\ref{eq:app-trig-scm-rf-coordinate} gives the TrigFlow target expressed in the RF output coordinate:
\[
    \mathbf T_{\mathrm{trig}\rightarrow\mathrm{RF}}
    =
    \frac{
    w_{\mathrm{trig}}Z^3\mathbf h_{\mathrm{RF}}
    }{
    w^2_{\mathrm{trig}}\|\mathbf h_{\mathrm{RF}}\|_2^2
    }
    =
    \frac{Z^3}{w_{\mathrm{trig}}}
    \frac{
    \mathbf h_{\mathrm{RF}}
    }{
    \|\mathbf h_{\mathrm{RF}}\|_2^2
    }.
\]
Thus, if \(w_{\mathrm{RF}}\) is kept general,
\[
    \mathbf T_{\mathrm{trig}\rightarrow\mathrm{RF}}
    =
    \frac{Z^3w_{\mathrm{RF}}}{w_{\mathrm{trig}}}\mathbf T_{\mathrm{RF}}.
\]
The loss gradient with respect to the RF velocity predictor satisfies
\begin{equation}
\label{eq:app-gradient-reweighting}
    \nabla_{\theta}\Lc_{\mathrm{TrigFlow\text{-}sCM}}
    =
    \frac{Zw_{\mathrm{RF}}}{w_{\mathrm{trig}}}
    \nabla_{\theta}\Lc_{\mathrm{RF\text{-}sCM}},
    \qquad (c=0).
\end{equation}
When \(c>0\), the difference is not reducible to a simple scalar reweighting, because the TrigFlow denominator becomes
$
    w^2_{\mathrm{trig}}\|\mathbf h_{\mathrm{RF}}\|_2^2+cZ^4
$, whereas the RF denominator is
$
    w_{\mathrm{RF}}^2\|\mathbf h_{\mathrm{RF}}\|_2^2+c
$. Therefore, tangent normalization breaks the strict equivalence of the two normalized MSE losses.

\paragraph{Finite-precision JVP evaluation.}
The relation in Eqn.~\ref{eq:app-scaled-rf-jvp} is exact only in real arithmetic. In floating-point computation,
\begin{equation}
\label{eq:app-jvp-finite-precision}
    \mathrm{fl}\!\left[
    \texttt{JVP}
    \left(
    \vv_{\theta^-};
    (\xv_u,u),
    b^2(\mathbf V,1)
    \right)
    \right]
    \neq
    b^2\,
    \mathrm{fl}\!\left[
    \texttt{JVP}
    \left(
    \vv_{\theta^-};
    (\xv_u,u),
    (\mathbf V,1)
    \right)
    \right]
\end{equation}
in general, and the JVP rearrangement~\citep{lu2024simplifying} further absorbs the coefficient $w_{\mathrm{trig}}(\tau)=\cos\tau$ into JVP computation. Placing the scales inside the JVP direction propagates the scaled tangent through every layer of the network, while factoring them outside first evaluates an unscaled tangent and only then rescales the result. These two evaluation orders can differ because of rounding, mixed-precision casts, fused kernels, activation checkpointing, and custom FlashAttention JVP implementations.

Moreover, the TrigFlow wrapper contains explicit input-output transform terms whose cancellations are algebraically exact but not necessarily bitwise exact. For example, the derivation of Eqn.~\ref{eq:app-trig-tangent-rf-tangent} cancels the state-dependent terms from differentiating \(C\xv_\tau\), \(S\Fv_{\theta^-}^{\mathrm{trig}}\), and the wrapper coefficients. A direct RF implementation computes the compact expression
\[
    \mathbf h_{\mathrm{RF}}
    =
    \mathbf V-\vv_{\theta^-}-u\mathbf J_{\theta^-}^{\mathrm{RF}}
\]
without these intermediate transform terms. Consequently, even when the exact-arithmetic tangent relation \(\mathbf h_{\mathrm{trig}}=b^2\mathbf h_{\mathrm{RF}}\) holds, the two implementations are not expected to be bitwise equivalent under practical large-scale mixed-precision training. This numerical distinction can be amplified by the normalized target \(\mathbf g/(\|\mathbf g\|_2^2+c)\), especially when \(\|\mathbf g\|_2\) is small or the stabilizing constant \(c\) is small.

%% file: sections/appendix_fa2_jvp_custom_mask.tex
\section{FlashAttention-2 JVP Kernel with Custom Masks}
\label{app:fa2-jvp-custom-mask}

For TF-sCM, the student network is evaluated on a packed sequence that concatenates clean context tokens and noisy target tokens under a TF attention mask. The Jacobian-vector-product (JVP) must be computed through exactly the same masked attention operator as the primal forward pass. A dense additive mask is conceptually simple but memory-inefficient for long video sequences. We therefore represent the custom mask as a sparse list of admissible query-key rectangles in the MagiAttention~\citep{magiattention2025} style, and stream only those rectangles inside the FlashAttention-2 loop.

Let \(\Mv\in\{0,-\infty\}^{N_q\times N_k}\) be a custom attention mask, where \(\Mv_{ab}=0\) means that query token \(a\) may attend to key token \(b\), and \(\Mv_{ab}=-\infty\) otherwise. The masked attention output is
\begin{equation}
    \vO
    =
    \softmax\!\left(\frac{\vQ\vK^\top}{\sqrt d}+\Mv\right)\vV .
\end{equation}
For JVP, given tangents \((\vtQ,\vtK,\vtV)\), the score tangent is
\begin{equation}
    \vtS
    =
    \frac{\vtQ\vK^\top+\vQ\vtK^\top}{\sqrt d}.
\end{equation}
The mask is a discrete routing object and has no tangent. Therefore, masked-out entries are assigned zero tangent contribution:
\[
    \vS_{ab}=-\infty,\quad \vtS_{ab}=0
    \qquad \text{if } \Mv_{ab}=-\infty .
\]
Equivalently, the JVP is taken through the masked attention map
\[
    (\vQ,\vK,\vV)\mapsto
    \softmax\!\left(\vS+\Mv\right)\vV,
\]
with \(\Mv\) fixed.

For a row \(a\), let \(p_{ab}\) denote the masked softmax probability over allowed keys. The attention tangent is
\begin{equation}
    \vtO_a
    =
    \sum_b p_{ab}\vtV_b
    +
    \sum_b p_{ab}
    \left(
    \vtS_{ab}
    -
    \sum_c p_{ac}\vtS_{ac}
    \right)
    \vV_b ,
\end{equation}
where all sums are over valid keys under \(\Mv\). The kernel computes this expression in the same online-softmax pass as the primal FlashAttention computation. For a streamed block, define the unnormalized probability
\[
    \tilde{\vP}_{ij}=\exp(\vS_{ij}-m_{\mathrm{new}}),
    \qquad
    \tilde{\vH}_{ij}=\tilde{\vP}_{ij}\odot \vtS_{ij}.
\]
Besides the standard FlashAttention accumulators \((m,\ell,\vO)\), we maintain three JVP accumulators:
\[
    \vA=\sum_j \tilde{\vP}_{ij}\vtV_j,\qquad
    \vB=\sum_j \tilde{\vH}_{ij}\vV_j,\qquad
    r=\sum_j \mathrm{rowsum}(\tilde{\vH}_{ij}).
\]
After normalization, the tangent output is
\begin{equation}
\label{eq:app-fa2-jvp-epilogue}
    \vtO_i
    =
    \diag(\ell_i)^{-1}
    \left(
    \vA_i+\vB_i-\diag(r_i)\vO_i
    \right),
\end{equation}
where \(\vO_i\) in the last term is the normalized primal output. The same online rescaling factor used for the primal accumulators is applied to \(\vA_i,\vB_i,r_i\), so the JVP remains numerically aligned with the FlashAttention-2 softmax normalization.

\paragraph{Sparse custom-mask representation.}
The custom mask is represented as a set of query groups and their admissible key ranges. Each query group \(g\) contains a contiguous query interval \(\mathcal Q_g=[q_g^0,q_g^1)\) and a list of valid key intervals
\[
    \mathcal K_g
    =
    \left\{
    [k_{g,r}^0,k_{g,r}^1)
    \right\}_{r=1}^{R_g}.
\]
The kernel launches tasks \((g,i)\), where \(i\) is a query tile inside \(\mathcal Q_g\). Each task streams only the key ranges in \(\mathcal K_g\). This range-list view covers both dense/full attention and structured causal masks: dense attention has one query group with one full key range, while teacher-forcing or block-causal masks are decomposed into a small number of full query-key rectangles. Importantly, the same sparse schedule is used for both the primal score \(\vS\) and its tangent \(\vtS\), ensuring that the tangent corresponds to the exact masked attention operator used in the forward pass.

We present the full algorithm in Algo.~\ref{algo:jvp-custom-mask}.
\begin{algorithm}[htbp]
  \caption{\label{algo:jvp-custom-mask} FlashAttention-2 Forward Pass with JVP Computation and Custom Mask}
  \begin{algorithmic}[1]
    \REQUIRE Matrices $\vQ,\vK,\vV$, their tangents $\vtQ,\vtK,\vtV$, block sizes $B_r,B_c$, and a custom mask $\Mv$ represented by query groups $\{\mathcal Q_g\}$, key ranges $\{\mathcal K_g\}$, and task list $\mathcal T=\{(g,i)\}$.
    \STATE Split $\vQ,\vtQ$ into query tiles of size $B_r\times d$ and $\vK,\vtK,\vV,\vtV$ into key/value tiles of size $B_c\times d$.
    \STATE Allocate output $\vO$, log-sum-exp $L$, and output tangent $\vtO$.
    \FOR{each task $(g,i)\in\mathcal T$ \textbf{in parallel}}
      \STATE Let $\mathcal I_i$ be the query-token indices of tile $i$ and let $\mathcal I_i^{g}=\mathcal I_i\cap\mathcal Q_g$.
      \STATE Load $\vQ_i,\vtQ_i$ from HBM to SRAM, with rows outside $\mathcal I_i^{g}$ masked out.
      \STATE Initialize $m_i\leftarrow(-\infty)^{B_r}$, $\ell_i\leftarrow\mathbf 0^{B_r}$, $\vO_i\leftarrow\mathbf 0^{B_r\times d}$.
      \STATE Initialize JVP accumulators $r_i\leftarrow\mathbf 0^{B_r}$, $\vA_i\leftarrow\mathbf 0^{B_r\times d}$, $\vB_i\leftarrow\mathbf 0^{B_r\times d}$.
      \FOR{each allowed key range $[k_0,k_1)\in\mathcal K_g$}
        \FOR{each key/value tile $j$ intersecting $[k_0,k_1)$}
          \STATE Load $\vK_j,\vtK_j,\vV_j,\vtV_j$ from HBM to SRAM.
          \STATE Let $\mathcal J_j$ be the key-token indices of tile $j$ and define the tile-valid mask
          \[
              \mathcal B_{ij}^{g}
              =
              \left\{
              (a,b): a\in\mathcal I_i^{g},\ b\in\mathcal J_j\cap[k_0,k_1)
              \right\}.
          \]
          \STATE Compute scores and score tangents $
              \vS_{ij}=\vQ_i\vK_j^\top,
              \vtS_{ij}=\vtQ_i\vK_j^\top+\vQ_i\vtK_j^\top .
          $
          \STATE Apply the custom mask 
          $
              \vS_{ij}\leftarrow\mathrm{where}(\mathcal B_{ij}^{g},\vS_{ij},-\infty),
              \vtS_{ij}\leftarrow\mathrm{where}(\mathcal B_{ij}^{g},\vtS_{ij},0).
          $
          \STATE Compute $m_{\mathrm{new}}=\max(m_i,\mathrm{rowmax}(\vS_{ij}))$.
          \STATE Compute $\tilde{\vP}_{ij}=\exp(\vS_{ij}-m_{\mathrm{new}})$.
          \STATE Compute $\ell_{\mathrm{new}}=e^{m_i-m_{\mathrm{new}}}\cdot\ell_i+\mathrm{rowsum}(\tilde{\vP}_{ij})$.
          \STATE Rescale primal and JVP accumulators:
          \[
          \begin{aligned}
              \vO_i &\leftarrow \diag(e^{m_i-m_{\mathrm{new}}})\vO_i,\\
              \vA_i &\leftarrow \diag(e^{m_i-m_{\mathrm{new}}})\vA_i,\\
              \vB_i &\leftarrow \diag(e^{m_i-m_{\mathrm{new}}})\vB_i,\\
              r_i &\leftarrow e^{m_i-m_{\mathrm{new}}}\cdot r_i .
          \end{aligned}
          \]
          \STATE Update primal accumulator $
              \vO_i\leftarrow \vO_i+\tilde{\vP}_{ij}\vV_j.
          $
          \STATE Update value-tangent accumulator $
              \vA_i\leftarrow \vA_i+\tilde{\vP}_{ij}\vtV_j.
          $
          \STATE Compute $\tilde{\vH}_{ij}=\tilde{\vP}_{ij}\odot\vtS_{ij}$.
          \STATE Update score-tangent accumulators:
          \[
          \begin{aligned}
              r_i &\leftarrow r_i+\mathrm{rowsum}(\tilde{\vH}_{ij}),\\
              \vB_i &\leftarrow \vB_i+\tilde{\vH}_{ij}\vV_j .
          \end{aligned}
          \]
          \STATE Update $m_i\leftarrow m_{\mathrm{new}}$, $\ell_i\leftarrow\ell_{\mathrm{new}}$.
        \ENDFOR
      \ENDFOR
      \STATE Normalize primal output:
      \[
          \vO_i\leftarrow \diag(\ell_i)^{-1}\vO_i,\qquad
          L_i\leftarrow m_i+\log(\ell_i).
      \]
      \STATE Compute the JVP epilogue:
      \[
          \vtO_i
          =
          \diag(\ell_i)^{-1}
          \left(
          \vA_i+\vB_i-\diag(r_i)\vO_i
          \right).
      \]
      \STATE Write $\vO_i,L_i,\vtO_i$ to HBM for rows in $\mathcal I_i^{g}$.
    \ENDFOR
    \RETURN $\vO,L,\vtO$.
  \end{algorithmic}
\end{algorithm}

%% file: main.bbl
\begin{thebibliography}{87}
\providecommand{\natexlab}[1]{#1}
\providecommand{\url}[1]{\texttt{#1}}
\expandafter\ifx\csname urlstyle\endcsname\relax
  \providecommand{\doi}[1]{doi: #1}\else
  \providecommand{\doi}{doi: \begingroup \urlstyle{rm}\Url}\fi

\bibitem[Ali et~al.(2025)Ali, Bai, Bala, Balaji, Blakeman, Cai, Cao, Cao, Cha, Chao, et~al.]{ali2025world}
Arslan Ali, Junjie Bai, Maciej Bala, Yogesh Balaji, Aaron Blakeman, Tiffany Cai, Jiaxin Cao, Tianshi Cao, Elizabeth Cha, Yu-Wei Chao, et~al.
\newblock World simulation with video foundation models for physical ai.
\newblock \emph{arXiv preprint arXiv:2511.00062}, 2025.

\bibitem[Arriola et~al.(2025)Arriola, Gokaslan, Chiu, Yang, Qi, Han, Sahoo, and Kuleshov]{arriola2025block}
Marianne Arriola, Aaron Gokaslan, Justin~T. Chiu, Zhihan Yang, Zhixuan Qi, Jiaqi Han, Subham~Sekhar Sahoo, and Volodymyr Kuleshov.
\newblock Block diffusion: Interpolating between autoregressive and diffusion language models.
\newblock \emph{arXiv preprint arXiv:2503.09573}, 2025.

\bibitem[Bao et~al.(2024)Bao, Xiang, Yue, He, Zhu, Zheng, Zhao, Liu, Wang, and Zhu]{bao2024vidu}
Fan Bao, Chendong Xiang, Gang Yue, Guande He, Hongzhou Zhu, Kaiwen Zheng, Min Zhao, Shilong Liu, Yaole Wang, and Jun Zhu.
\newblock Vidu: a highly consistent, dynamic and skilled text-to-video generator with diffusion models.
\newblock \emph{arXiv preprint arXiv:2405.04233}, 2024.

\bibitem[Brooks et~al.(2024)Brooks, Peebles, Holmes, DePue, Guo, Jing, Schnurr, Taylor, Luhman, Luhman, et~al.]{brooks2024video}
Tim Brooks, Bill Peebles, Connor Holmes, Will DePue, Yufei Guo, Li~Jing, David Schnurr, Joe Taylor, Troy Luhman, Eric Luhman, et~al.
\newblock Video generation models as world simulators. 2024.
\newblock \emph{URL https://openai. com/research/video-generation-models-as-world-simulators}, 3, 2024.

\bibitem[Cai et~al.(2026)Cai, Nie, Liu, Berner, Zhang, Ma, Chen, Agrawala, Guibas, Wetzstein, et~al.]{cai2026mode}
Shengqu Cai, Weili Nie, Chao Liu, Julius Berner, Lvmin Zhang, Nanye Ma, Hansheng Chen, Maneesh Agrawala, Leonidas Guibas, Gordon Wetzstein, et~al.
\newblock Mode seeking meets mean seeking for fast long video generation.
\newblock \emph{arXiv preprint arXiv:2602.24289}, 2026.

\bibitem[Chen et~al.(2024)Chen, Mart{\'\i}~Mons{\'o}, Du, Simchowitz, Tedrake, and Sitzmann]{chen2024diffusion}
Boyuan Chen, Diego Mart{\'\i}~Mons{\'o}, Yilun Du, Max Simchowitz, Russ Tedrake, and Vincent Sitzmann.
\newblock Diffusion forcing: Next-token prediction meets full-sequence diffusion.
\newblock \emph{Advances in Neural Information Processing Systems}, 37:\penalty0 24081--24125, 2024.

\bibitem[Chen et~al.(2025{\natexlab{a}})Chen, Lin, Yang, Lin, Zhu, Fan, Zhang, Chen, Chen, Ma, et~al.]{chen2025skyreels}
Guibin Chen, Dixuan Lin, Jiangping Yang, Chunze Lin, Junchen Zhu, Mingyuan Fan, Hao Zhang, Sheng Chen, Zheng Chen, Chengcheng Ma, et~al.
\newblock Skyreels-v2: Infinite-length film generative model.
\newblock \emph{arXiv preprint arXiv:2504.13074}, 2025{\natexlab{a}}.

\bibitem[Chen et~al.(2025{\natexlab{b}})Chen, Zhang, Tan, Guibas, Wetzstein, and Bi]{chen2025pi}
Hansheng Chen, Kai Zhang, Hao Tan, Leonidas Guibas, Gordon Wetzstein, and Sai Bi.
\newblock pi-flow: Policy-based few-step generation via imitation distillation.
\newblock \emph{arXiv preprint arXiv:2510.14974}, 2025{\natexlab{b}}.

\bibitem[Chen et~al.(2026)Chen, Wang, Huang, Yang, Zhang, Xiao, Chu, Mao, Hu, Liu, Zhao, Mao, Chen, Xie, Qi, and Han]{longlive_2.0}
Yukang Chen, Luozhou Wang, Wei Huang, Shuai Yang, Bohan Zhang, Yicheng Xiao, Ruihang Chu, Weian Mao, Qixin Hu, Shaoteng Liu, Yuyang Zhao, Huizi Mao, Ying-Cong Chen, Enze Xie, Xiaojuan Qi, and Song Han.
\newblock Longlive2.0: An nvfp4 parallel infrastructure for long video generation.
\newblock \emph{arXiv preprint arXiv}, 2026.

\bibitem[Cheng et~al.(2025)Cheng, Sun, Li, and Lin]{cheng2025twinflow}
Zhenglin Cheng, Peng Sun, Jianguo Li, and Tao Lin.
\newblock Twinflow: Realizing one-step generation on large models with self-adversarial flows.
\newblock \emph{arXiv preprint arXiv:2512.05150}, 2025.

\bibitem[Dong et~al.(2024)Dong, Feng, Guessous, Liang, and He]{dong2024flex}
Juechu Dong, Boyuan Feng, Driss Guessous, Yanbo Liang, and Horace He.
\newblock Flex attention: A programming model for generating optimized attention kernels.
\newblock \emph{arXiv preprint arXiv:2412.05496}, 2\penalty0 (3):\penalty0 4, 2024.

\bibitem[Feng et~al.(2026)Feng, Cui, Ban, and Hsieh]{feng2026one}
Jiaqi Feng, Justin Cui, Yuanhao Ban, and Cho-Jui Hsieh.
\newblock One-forcing: Towards stable one-step autoregressive video generation.
\newblock \emph{arXiv preprint arXiv:2605.23458}, 2026.

\bibitem[Feng et~al.(2025)Feng, Xiang, Mao, Tan, Zhang, Huang, Zheng, Liu, Su, and Zhu]{feng2025vidarc}
Yao Feng, Chendong Xiang, Xinyi Mao, Hengkai Tan, Zuyue Zhang, Shuhe Huang, Kaiwen Zheng, Haitian Liu, Hang Su, and Jun Zhu.
\newblock Vidarc: Embodied video diffusion model for closed-loop control.
\newblock \emph{arXiv preprint arXiv:2512.17661}, 2025.

\bibitem[Gao et~al.(2025)Gao, Guo, Hoang, Huang, Jiang, Kong, Li, Li, Li, Li, et~al.]{gao2025seedance}
Yu~Gao, Haoyuan Guo, Tuyen Hoang, Weilin Huang, Lu~Jiang, Fangyuan Kong, Huixia Li, Jiashi Li, Liang Li, Xiaojie Li, et~al.
\newblock Seedance 1.0: Exploring the boundaries of video generation models.
\newblock \emph{arXiv preprint arXiv:2506.09113}, 2025.

\bibitem[Geng et~al.(2025)Geng, Deng, Bai, Kolter, and He]{geng2025mean}
Zhengyang Geng, Mingyang Deng, Xingjian Bai, J~Zico Kolter, and Kaiming He.
\newblock Mean flows for one-step generative modeling.
\newblock \emph{arXiv preprint arXiv:2505.13447}, 2025.

\bibitem[Gu et~al.(2026)Gu, Fang, Jiang, Mao, Han, Cai, and Shou]{gu2026anyflow}
Yuchao Gu, Guian Fang, Yuxin Jiang, Weijia Mao, Song Han, Han Cai, and Mike~Zheng Shou.
\newblock Anyflow: Any-step video diffusion model with on-policy flow map distillation.
\newblock \emph{arXiv preprint arXiv:2605.13724}, 2026.

\bibitem[{Hao-AI Lab}(2026)]{fastvideo_codebase}
{Hao-AI Lab}.
\newblock {FastVideo}: A unified inference and post-training framework for accelerated video generation, 2026.
\newblock URL \url{https://github.com/hao-ai-lab/FastVideo}.

\bibitem[He et~al.(2025)He, Peng, Liu, Wang, Zhang, Cui, Kang, Jiang, An, Ren, Xu, Guo, Gong, Wu, Li, Song, Liu, Li, and Zhou]{he2025matrixgame}
Xianglong He, Chunli Peng, Zexiang Liu, Boyang Wang, Yifan Zhang, Qi~Cui, Fei Kang, Biao Jiang, Mengyin An, Yangyang Ren, Baixin Xu, Hao-Xiang Guo, Kaixiong Gong, Size Wu, Wei Li, Xuchen Song, Yang Liu, Yangguang Li, and Yahui Zhou.
\newblock Matrix-game 2.0: An open-source real-time and streaming interactive world model.
\newblock \emph{arXiv preprint arXiv:2508.13009}, 2025.

\bibitem[Ho et~al.(2020)Ho, Jain, and Abbeel]{ho2020denoising}
Jonathan Ho, Ajay Jain, and Pieter Abbeel.
\newblock Denoising diffusion probabilistic models.
\newblock \emph{Advances in neural information processing systems}, 33:\penalty0 6840--6851, 2020.

\bibitem[Hong et~al.(2025)Hong, Mei, Ge, Xu, Zhou, Bi, Hold-Geoffroy, Roberts, Fisher, Shechtman, Sunkavalli, Liu, Li, and Tan]{hong2025relic}
Yicong Hong, Yiqun Mei, Chongjian Ge, Yiran Xu, Yang Zhou, Sai Bi, Yannick Hold-Geoffroy, Mike Roberts, Matthew Fisher, Eli Shechtman, Kalyan Sunkavalli, Feng Liu, Zhengqi Li, and Hao Tan.
\newblock Relic: Interactive video world model with long-horizon memory.
\newblock \emph{arXiv preprint arXiv:2512.04040}, 2025.

\bibitem[Hoogeboom et~al.(2023)Hoogeboom, Heek, and Salimans]{hoogeboom2023simple}
Emiel Hoogeboom, Jonathan Heek, and Tim Salimans.
\newblock simple diffusion: End-to-end diffusion for high resolution images.
\newblock In \emph{International Conference on Machine Learning}, pages 13213--13232. PMLR, 2023.

\bibitem[Huang et~al.(2025{\natexlab{a}})Huang, Li, He, Zhou, and Shechtman]{huang2025selfforcing}
Xun Huang, Zhengqi Li, Guande He, Mingyuan Zhou, and Eli Shechtman.
\newblock Self forcing: Bridging the train-test gap in autoregressive video diffusion.
\newblock \emph{arXiv preprint arXiv:2506.08009}, 2025{\natexlab{a}}.

\bibitem[Huang et~al.(2025{\natexlab{b}})Huang, Guo, Wu, Wang, Zhang, Huang, Gan, Liu, Zhao, Chen, Liu, and Hoi]{huang2025live}
Yubo Huang, Hailong Guo, Fangtai Wu, Weiqiang Wang, Shifeng Zhang, Shijie Huang, Qijun Gan, Lin Liu, Sirui Zhao, Enhong Chen, Jiaming Liu, and Steven Hoi.
\newblock Live avatar: Streaming real-time audio-driven avatar generation with infinite length.
\newblock \emph{arXiv preprint arXiv:2512.04677}, 2025{\natexlab{b}}.

\bibitem[Huang et~al.(2024)Huang, He, Yu, Zhang, Si, Jiang, Zhang, Wu, Jin, Chanpaisit, et~al.]{huang2024vbench}
Ziqi Huang, Yinan He, Jiashuo Yu, Fan Zhang, Chenyang Si, Yuming Jiang, Yuanhan Zhang, Tianxing Wu, Qingyang Jin, Nattapol Chanpaisit, et~al.
\newblock Vbench: Comprehensive benchmark suite for video generative models.
\newblock In \emph{Proceedings of the IEEE/CVF Conference on Computer Vision and Pattern Recognition}, pages 21807--21818, 2024.

\bibitem[HunyuanWorld(2025)]{hunyuanworld2025hy}
Team HunyuanWorld.
\newblock Hy-world 1.5: A systematic framework for interactive world modeling with real-time latency and geometric consistency.
\newblock \emph{arXiv preprint}, 2025.

\bibitem[Jacobs et~al.(2023)Jacobs, Tanaka, Zhang, Zhang, Song, Rajbhandari, and He]{jacobs2023deepspeed}
Sam~Ade Jacobs, Masahiro Tanaka, Chengming Zhang, Minjia Zhang, Shuaiwen~Leon Song, Samyam Rajbhandari, and Yuxiong He.
\newblock Deepspeed ulysses: System optimizations for enabling training of extreme long sequence transformer models.
\newblock \emph{arXiv preprint arXiv:2309.14509}, 2023.

\bibitem[Jiang et~al.(2025)Jiang, Liu, Wang, Wu, Li, Li, Jin, Liu, Lu, Li, et~al.]{jiang2025distribution}
Dengyang Jiang, Dongyang Liu, Zanyi Wang, Qilong Wu, Liuzhuozheng Li, Hengzhuang Li, Xin Jin, David Liu, Changsheng Lu, Zhen Li, et~al.
\newblock Distribution matching distillation meets reinforcement learning.
\newblock \emph{arXiv preprint arXiv:2511.13649}, 2025.

\bibitem[Jin et~al.(2025)Jin, Sun, Li, Xu, Jiang, Zhuang, Huang, Song, Mu, and Lin]{jin2025pyramidal}
Yang Jin, Zhicheng Sun, Ningyuan Li, Kun Xu, Hao Jiang, Nan Zhuang, Quzhe Huang, Yang Song, Yadong Mu, and Zhouchen Lin.
\newblock Pyramidal flow matching for efficient video generative modeling.
\newblock In \emph{International Conference on Learning Representations}, volume 2025, pages 23378--23402, 2025.

\bibitem[Kim et~al.(2023)Kim, Lai, Liao, Murata, Takida, Uesaka, He, Mitsufuji, and Ermon]{kim2023consistency}
Dongjun Kim, Chieh-Hsin Lai, Wei-Hsiang Liao, Naoki Murata, Yuhta Takida, Toshimitsu Uesaka, Yutong He, Yuki Mitsufuji, and Stefano Ermon.
\newblock Consistency trajectory models: Learning probability flow ode trajectory of diffusion.
\newblock \emph{arXiv preprint arXiv:2310.02279}, 2023.

\bibitem[Kim et~al.(2026)Kim, Hu, Kuo, and Beerel]{kim2026memrope}
Youngrae Kim, Qixin Hu, C-C~Jay Kuo, and Peter~A Beerel.
\newblock Memrope: Training-free infinite video generation via evolving memory tokens.
\newblock \emph{arXiv preprint arXiv:2603.12513}, 2026.

\bibitem[Kong et~al.(2024)Kong, Tian, Zhang, Min, Dai, Zhou, Xiong, Li, Wu, Zhang, et~al.]{kong2024hunyuanvideo}
Weijie Kong, Qi~Tian, Zijian Zhang, Rox Min, Zuozhuo Dai, Jin Zhou, Jiangfeng Xiong, Xin Li, Bo~Wu, Jianwei Zhang, et~al.
\newblock Hunyuanvideo: A systematic framework for large video generative models.
\newblock \emph{arXiv preprint arXiv:2412.03603}, 2024.

\bibitem[Li et~al.(2026{\natexlab{a}})Li, Zeng, Lu, Zhu, Ouyang, Wang, Cheng, Shen, and Zhang]{li2026aad}
Haobo Li, Yanhong Zeng, Yunhong Lu, Jiapeng Zhu, Hao Ouyang, Qiuyu Wang, Ka~Leong Cheng, Yujun Shen, and Zhipeng Zhang.
\newblock Aad-1: Asymmetric adversarial distillation for one-step autoregressive video generation.
\newblock \emph{arXiv preprint arXiv:2606.03972}, 2026{\natexlab{a}}.

\bibitem[Li et~al.(2026{\natexlab{b}})Li, Liu, Lin, and Chandraker]{li2026rolling}
Haodong Li, Shaoteng Liu, Zhe Lin, and Manmohan Chandraker.
\newblock Rolling sink: Bridging limited-horizon training and open-ended testing in autoregressive video diffusion.
\newblock \emph{arXiv preprint arXiv:2602.07775}, 2026{\natexlab{b}}.

\bibitem[Li et~al.(2026{\natexlab{c}})Li, Zhang, Luo, Yang, Wang, Han, Yu, Gao, Xue, Zhu, et~al.]{li2026causal}
Lin Li, Qihang Zhang, Yiming Luo, Shuai Yang, Ruilin Wang, Fei Han, Mingrui Yu, Zelin Gao, Nan Xue, Xing Zhu, et~al.
\newblock Causal world modeling for robot control.
\newblock \emph{arXiv preprint arXiv:2601.21998}, 2026{\natexlab{c}}.

\bibitem[Lin et~al.(2025{\natexlab{a}})Lin, Xia, Ren, Yang, Xiao, and Jiang]{lin2025diffusion}
Shanchuan Lin, Xin Xia, Yuxi Ren, Ceyuan Yang, Xuefeng Xiao, and Lu~Jiang.
\newblock Diffusion adversarial post-training for one-step video generation.
\newblock In \emph{Proceedings of the 42nd International Conference on Machine Learning}, volume 267 of \emph{Proceedings of Machine Learning Research}, pages 37959--37974. PMLR, 2025{\natexlab{a}}.

\bibitem[Lin et~al.(2025{\natexlab{b}})Lin, Yang, He, Jiang, Ren, Xia, Zhao, Xiao, and Jiang]{lin2025autoregressive}
Shanchuan Lin, Ceyuan Yang, Hao He, Jianwen Jiang, Yuxi Ren, Xin Xia, Yang Zhao, Xuefeng Xiao, and Lu~Jiang.
\newblock Autoregressive adversarial post-training for real-time interactive video generation.
\newblock \emph{arXiv preprint arXiv:2506.09350}, 2025{\natexlab{b}}.

\bibitem[Lin et~al.(2026)Lin, Yang, Lin, Chen, and Fan]{lin2026continuous}
Shanchuan Lin, Ceyuan Yang, Zhijie Lin, Hao Chen, and Haoqi Fan.
\newblock Continuous adversarial flow models.
\newblock \emph{arXiv preprint arXiv:2604.11521}, 2026.

\bibitem[Lipman et~al.(2022)Lipman, Chen, Ben-Hamu, Nickel, and Le]{lipman2022flow}
Yaron Lipman, Ricky~TQ Chen, Heli Ben-Hamu, Maximilian Nickel, and Matt Le.
\newblock Flow matching for generative modeling.
\newblock \emph{arXiv preprint arXiv:2210.02747}, 2022.

\bibitem[Liu et~al.(2026)Liu, Liu, Mei, Wen, Yang, and Liu]{liu2026diagdistill}
Jinxiu Liu, Xuanming Liu, Kangfu Mei, Yandong Wen, Ming-Hsuan Yang, and Weiyang Liu.
\newblock Streaming autoregressive video generation via diagonal distillation.
\newblock In \emph{ICLR}, 2026.

\bibitem[Liu et~al.(2022)Liu, Gong, and Liu]{liu2022flow}
Xingchao Liu, Chengyue Gong, and Qiang Liu.
\newblock Flow straight and fast: Learning to generate and transfer data with rectified flow.
\newblock \emph{arXiv preprint arXiv:2209.03003}, 2022.

\bibitem[Lu and Song(2024)]{lu2024simplifying}
Cheng Lu and Yang Song.
\newblock Simplifying, stabilizing and scaling continuous-time consistency models.
\newblock \emph{arXiv preprint arXiv:2410.11081}, 2024.

\bibitem[Lu et~al.(2022)Lu, Zheng, Bao, Chen, Li, and Zhu]{lu2022maximum}
Cheng Lu, Kaiwen Zheng, Fan Bao, Jianfei Chen, Chongxuan Li, and Jun Zhu.
\newblock Maximum likelihood training for score-based diffusion odes by high order denoising score matching.
\newblock In \emph{International conference on machine learning}, pages 14429--14460. PMLR, 2022.

\bibitem[Luhman and Luhman(2021)]{luhman2021knowledge}
Eric Luhman and Troy Luhman.
\newblock Knowledge distillation in iterative generative models for improved sampling speed.
\newblock \emph{arXiv preprint arXiv:2101.02388}, 2021.

\bibitem[Nie et~al.(2026{\natexlab{a}})Nie, Berner, Liu, and Vahdat]{fastgen2026}
Weili Nie, Julius Berner, Chao Liu, and Arash Vahdat.
\newblock Nvidia fastgen: Fast generation from diffusion models, 2026{\natexlab{a}}.
\newblock URL \url{https://github.com/NVlabs/FastGen}.

\bibitem[Nie et~al.(2026{\natexlab{b}})Nie, Berner, Ma, Liu, Xie, and Vahdat]{nie2026transition}
Weili Nie, Julius Berner, Nanye Ma, Chao Liu, Saining Xie, and Arash Vahdat.
\newblock Transition matching distillation for fast video generation.
\newblock \emph{arXiv preprint arXiv:2601.09881}, 2026{\natexlab{b}}.

\bibitem[Ning et~al.(2024)Ning, Li, Su, Salah, and Onal~Ertugrul]{ning2024elucidating}
Mang Ning, Mingxiao Li, Jianlin Su, Albert~Ali Salah, and Itir Onal~Ertugrul.
\newblock Elucidating the exposure bias in diffusion models.
\newblock In \emph{International Conference on Learning Representations}, volume 2024, pages 15167--15189, 2024.

\bibitem[{NVIDIA}(2026)]{nvidia2026cosmos3}
{NVIDIA}.
\newblock Cosmos 3: Omnimodal world models for physical ai.
\newblock \emph{arXiv preprint arXiv:2606.02800}, 2026.
\newblock URL \url{https://research.nvidia.com/labs/cosmos-lab/cosmos3/technical-report.pdf}.

\bibitem[Park et~al.(2026)Park, Li, Tulyakov, and Kag]{park2026eflow}
Dogyun Park, Yanyu Li, Sergey Tulyakov, and Anil Kag.
\newblock Eflow: Fast few-step video generator training from scratch via efficient solution flow.
\newblock \emph{arXiv preprint arXiv:2603.27086}, 2026.

\bibitem[Peng et~al.(2025)Peng, Zhu, Liu, Wu, Li, Sun, and Wu]{peng2025facm}
Yansong Peng, Kai Zhu, Yu~Liu, Pingyu Wu, Hebei Li, Xiaoyan Sun, and Feng Wu.
\newblock Facm: Flow-anchored consistency models.
\newblock \emph{arXiv preprint arXiv:2507.03738}, 2025.

\bibitem[Pu et~al.(2025)Pu, Han, Tang, Tang, Wang, Zhuang, and Huang]{pu2025few}
Yifan Pu, Yizeng Han, Zhiwei Tang, Jiasheng Tang, Fan Wang, Bohan Zhuang, and Gao Huang.
\newblock Few-step distillation for text-to-image generation: A practical guide.
\newblock \emph{arXiv preprint arXiv:2512.13006}, 2025.

\bibitem[{Robbyant Team} et~al.(2026){Robbyant Team}, Gao, Wang, Zeng, Zhu, Cheng, Li, Wang, Xu, Ma, Chen, Liu, Cheng, Yao, Zhu, Meng, Zheng, Bai, Chen, Shen, Yu, Zhu, Shen, and Ouyang]{gao2026advancing}
{Robbyant Team}, Zelin Gao, Qiuyu Wang, Yanhong Zeng, Jiapeng Zhu, Ka~Leong Cheng, Yixuan Li, Hanlin Wang, Yinghao Xu, Shuailei Ma, Yihang Chen, Jie Liu, Yansong Cheng, Yao Yao, Jiayi Zhu, Yihao Meng, Kecheng Zheng, Qingyan Bai, Jingye Chen, Zehong Shen, Yue Yu, Xing Zhu, Yujun Shen, and Hao Ouyang.
\newblock Advancing open-source world models.
\newblock \emph{arXiv preprint arXiv:2601.20540}, 2026.

\bibitem[Sabour et~al.(2025)Sabour, Fidler, and Kreis]{sabour2025align}
Amirmojtaba Sabour, Sanja Fidler, and Karsten Kreis.
\newblock Align your flow: Scaling continuous-time flow map distillation.
\newblock \emph{arXiv preprint arXiv:2506.14603}, 2025.

\bibitem[Sahoo et~al.(2024)Sahoo, Arriola, Schiff, Gokaslan, Marroquin, Chiu, Rush, and Kuleshov]{sahoo2024simple}
Subham~Sekhar Sahoo, Marianne Arriola, Yair Schiff, Aaron Gokaslan, Edgar Marroquin, Justin~T Chiu, Alexander Rush, and Volodymyr Kuleshov.
\newblock Simple and effective masked diffusion language models.
\newblock \emph{arXiv preprint arXiv:2406.07524}, 2024.

\bibitem[Schmidt(2019)]{schmidt2019generalization}
Florian Schmidt.
\newblock Generalization in generation: A closer look at exposure bias.
\newblock In \emph{Proceedings of the 3rd Workshop on Neural Generation and Translation}, pages 157--167, 2019.

\bibitem[Seedance et~al.(2026)Seedance, Chen, Chen, Chen, Chen, Chen, Chen, Cheng, Cheng, Cheng, et~al.]{seedance2026seedance}
Team Seedance, De~Chen, Liyang Chen, Xin Chen, Ying Chen, Zhuo Chen, Zhuowei Chen, Feng Cheng, Tianheng Cheng, Yufeng Cheng, et~al.
\newblock Seedance 2.0: Advancing video generation for world complexity.
\newblock \emph{arXiv preprint arXiv:2604.14148}, 2026.

\bibitem[Shi et~al.(2024)Shi, Han, Wang, Doucet, and Titsias]{shi2024simplified}
Jiaxin Shi, Kehang Han, Zhe Wang, Arnaud Doucet, and Michalis~K Titsias.
\newblock Simplified and generalized masked diffusion for discrete data.
\newblock \emph{arXiv preprint arXiv:2406.04329}, 2024.

\bibitem[Song et~al.(2020)Song, Sohl-Dickstein, Kingma, Kumar, Ermon, and Poole]{song2020score}
Yang Song, Jascha Sohl-Dickstein, Diederik~P Kingma, Abhishek Kumar, Stefano Ermon, and Ben Poole.
\newblock Score-based generative modeling through stochastic differential equations.
\newblock \emph{arXiv preprint arXiv:2011.13456}, 2020.

\bibitem[Song et~al.(2023)Song, Dhariwal, Chen, and Sutskever]{song2023consistency}
Yang Song, Prafulla Dhariwal, Mark Chen, and Ilya Sutskever.
\newblock Consistency models.
\newblock In \emph{International Conference on Machine Learning}, pages 32211--32252. PMLR, 2023.

\bibitem[Teng et~al.(2025)Teng, Jia, Sun, Li, Li, Tang, Han, Zhang, Zhang, Luo, et~al.]{teng2025magi}
Hansi Teng, Hongyu Jia, Lei Sun, Lingzhi Li, Maolin Li, Mingqiu Tang, Shuai Han, Tianning Zhang, WQ~Zhang, Weifeng Luo, et~al.
\newblock Magi-1: Autoregressive video generation at scale.
\newblock \emph{arXiv preprint arXiv:2505.13211}, 2025.

\bibitem[Tong et~al.(2025)Tong, Ma, Xie, and Jaakkola]{tong2025flow}
Shangyuan Tong, Nanye Ma, Saining Xie, and Tommi Jaakkola.
\newblock Flow map distillation without data.
\newblock \emph{arXiv preprint arXiv:2511.19428}, 2025.

\bibitem[Wan et~al.(2025)Wan, Wang, Ai, Wen, Mao, Xie, Chen, Yu, Zhao, Yang, Zeng, Wang, Zhang, Zhou, Wang, Chen, Zhu, Zhao, Yan, Huang, Feng, Zhang, Li, Wu, Chu, Feng, Zhang, Sun, Fang, Wang, Gui, Weng, Shen, Lin, Wang, Wang, Zhou, Wang, Shen, Yu, Shi, Huang, Xu, Kou, Lv, Li, Liu, Wang, Zhang, Huang, Li, Wu, Liu, Pan, Zheng, Hong, Shi, Feng, Jiang, Han, Wu, and Liu]{wan2025}
Team Wan, Ang Wang, Baole Ai, Bin Wen, Chaojie Mao, Chen-Wei Xie, Di~Chen, Feiwu Yu, Haiming Zhao, Jianxiao Yang, Jianyuan Zeng, Jiayu Wang, Jingfeng Zhang, Jingren Zhou, Jinkai Wang, Jixuan Chen, Kai Zhu, Kang Zhao, Keyu Yan, Lianghua Huang, Mengyang Feng, Ningyi Zhang, Pandeng Li, Pingyu Wu, Ruihang Chu, Ruili Feng, Shiwei Zhang, Siyang Sun, Tao Fang, Tianxing Wang, Tianyi Gui, Tingyu Weng, Tong Shen, Wei Lin, Wei Wang, Wei Wang, Wenmeng Zhou, Wente Wang, Wenting Shen, Wenyuan Yu, Xianzhong Shi, Xiaoming Huang, Xin Xu, Yan Kou, Yangyu Lv, Yifei Li, Yijing Liu, Yiming Wang, Yingya Zhang, Yitong Huang, Yong Li, You Wu, Yu~Liu, Yulin Pan, Yun Zheng, Yuntao Hong, Yupeng Shi, Yutong Feng, Zeyinzi Jiang, Zhen Han, Zhi-Fan Wu, and Ziyu Liu.
\newblock Wan: Open and advanced large-scale video generative models.
\newblock \emph{arXiv preprint arXiv:2503.20314}, 2025.

\bibitem[Wang et~al.(2023)Wang, Lu, Wang, Bao, Li, Su, and Zhu]{wang2023prolificdreamer}
Zhengyi Wang, Cheng Lu, Yikai Wang, Fan Bao, Chongxuan Li, Hang Su, and Jun Zhu.
\newblock Prolificdreamer: High-fidelity and diverse text-to-3d generation with variational score distillation.
\newblock \emph{Advances in neural information processing systems}, 36:\penalty0 8406--8441, 2023.

\bibitem[Wu et~al.(2026)Wu, Li, Zhang, and Ma]{wu2026diversity}
Tianhe Wu, Ruibin Li, Lei Zhang, and Kede Ma.
\newblock Diversity-preserved distribution matching distillation for fast visual synthesis.
\newblock \emph{arXiv preprint arXiv:2602.03139}, 2026.

\bibitem[Yang et~al.(2026)Yang, Huang, Chu, Xiao, Zhao, Wang, Li, Xie, Chen, Lu, et~al.]{longlive}
Shuai Yang, Wei Huang, Ruihang Chu, Yicheng Xiao, Yuyang Zhao, Xianbang Wang, Muyang Li, Enze Xie, Yingcong Chen, Yao Lu, et~al.
\newblock Longlive: Real-time interactive long video generation.
\newblock In \emph{ICLR}, 2026.

\bibitem[Ye et~al.(2025)Ye, Zheng, Xu, Li, Chen, Han, Liu, Zhang, Mao, Hao, et~al.]{ye2025data}
Haotian Ye, Kaiwen Zheng, Jiashu Xu, Puheng Li, Huayu Chen, Jiaqi Han, Sheng Liu, Qinsheng Zhang, Hanzi Mao, Zekun Hao, et~al.
\newblock Data-regularized reinforcement learning for diffusion models at scale.
\newblock \emph{arXiv preprint arXiv:2512.04332}, 2025.

\bibitem[Ye et~al.(2026)Ye, Ge, Zheng, Gao, Yu, Kurian, Indupuru, Tan, Zhu, Xiang, et~al.]{ye2026world}
Seonghyeon Ye, Yunhao Ge, Kaiyuan Zheng, Shenyuan Gao, Sihyun Yu, George Kurian, Suneel Indupuru, You~Liang Tan, Chuning Zhu, Jiannan Xiang, et~al.
\newblock World action models are zero-shot policies.
\newblock \emph{arXiv preprint arXiv:2602.15922}, 2026.

\bibitem[Yesiltepe et~al.(2026)Yesiltepe, Meral, Akan, Oktay, and Yanardag]{yesiltepe2026infinity}
Hidir Yesiltepe, Tuna Meral, Adil~Kaan Akan, Kaan Oktay, and Pinar Yanardag.
\newblock Infinity-rope: Action-controllable infinite video generation emerges from autoregressive self-rollout.
\newblock In \emph{Proceedings of the IEEE/CVF Conference on Computer Vision and Pattern Recognition}, pages 40256--40265, 2026.

\bibitem[Yi et~al.(2025)Yi, Jang, Cho, Nam, Yoon, and Kim]{yi2025deep}
Jung Yi, Wooseok Jang, Paul~Hyunbin Cho, Jisu Nam, Heeji Yoon, and Seungryong Kim.
\newblock Deep forcing: Training-free long video generation with deep sink and participative compression.
\newblock \emph{arXiv preprint arXiv:2512.05081}, 2025.

\bibitem[Yin et~al.(2024{\natexlab{a}})Yin, Gharbi, Park, Zhang, Shechtman, Durand, and Freeman]{yin2024improved}
Tianwei Yin, Micha{\"e}l Gharbi, Taesung Park, Richard Zhang, Eli Shechtman, Fredo Durand, and Bill Freeman.
\newblock Improved distribution matching distillation for fast image synthesis.
\newblock \emph{Advances in neural information processing systems}, 37:\penalty0 47455--47487, 2024{\natexlab{a}}.

\bibitem[Yin et~al.(2024{\natexlab{b}})Yin, Gharbi, Zhang, Shechtman, Durand, Freeman, and Park]{yin2024one}
Tianwei Yin, Micha{\"e}l Gharbi, Richard Zhang, Eli Shechtman, Fredo Durand, William~T Freeman, and Taesung Park.
\newblock One-step diffusion with distribution matching distillation.
\newblock In \emph{Proceedings of the IEEE/CVF conference on computer vision and pattern recognition}, pages 6613--6623, 2024{\natexlab{b}}.

\bibitem[Yin et~al.(2025)Yin, Zhang, Zhang, Freeman, Durand, Shechtman, and Huang]{yin2025slow}
Tianwei Yin, Qiang Zhang, Richard Zhang, William~T Freeman, Fredo Durand, Eli Shechtman, and Xun Huang.
\newblock From slow bidirectional to fast autoregressive video diffusion models.
\newblock In \emph{Proceedings of the Computer Vision and Pattern Recognition Conference}, pages 22963--22974, 2025.

\bibitem[You et~al.(2026)You, Li, Li, Mu, Chen, and Jiang]{you2026adaptive}
Yuyang You, Yongzhi Li, Jiahui Li, Yadong Mu, Quan Chen, and Peng Jiang.
\newblock Adaptive video distillation: Mitigating oversaturation and temporal collapse in few-step generation.
\newblock \emph{arXiv preprint arXiv:2603.21864}, 2026.

\bibitem[Zewei and Yunpeng(2025)]{magiattention2025}
Tao Zewei and Huang Yunpeng.
\newblock Magiattention: A distributed attention towards linear scalability for ultra-long context, heterogeneous mask training.
\newblock \url{https://github.com/SandAI-org/MagiAttention/}, 2025.

\bibitem[Zhang et~al.(2025{\natexlab{a}})Zhang, Wang, Jiang, Yang, Zheng, Xi, Wang, Zhu, Zhao, Stoica, et~al.]{zhang2025sla}
Jintao Zhang, Haoxu Wang, Kai Jiang, Shuo Yang, Kaiwen Zheng, Haocheng Xi, Ziteng Wang, Hongzhou Zhu, Min Zhao, Ion Stoica, et~al.
\newblock Sla: Beyond sparsity in diffusion transformers via fine-tunable sparse-linear attention.
\newblock \emph{arXiv preprint arXiv:2509.24006}, 2025{\natexlab{a}}.

\bibitem[Zhang et~al.(2025{\natexlab{b}})Zhang, Zheng, Jiang, Wang, Stoica, Gonzalez, Chen, and Zhu]{zhang2025turbodiffusion}
Jintao Zhang, Kaiwen Zheng, Kai Jiang, Haoxu Wang, Ion Stoica, Joseph~E Gonzalez, Jianfei Chen, and Jun Zhu.
\newblock Turbodiffusion: Accelerating video diffusion models by 100-200 times.
\newblock \emph{arXiv preprint arXiv:2512.16093}, 2025{\natexlab{b}}.

\bibitem[Zhang et~al.(2026)Zhang, Wang, Jiang, Zheng, Jiang, Stoica, Chen, Zhu, and Gonzalez]{zhang2026sla2}
Jintao Zhang, Haoxu Wang, Kai Jiang, Kaiwen Zheng, Youhe Jiang, Ion Stoica, Jianfei Chen, Jun Zhu, and Joseph~E Gonzalez.
\newblock Sla2: Sparse-linear attention with learnable routing and qat.
\newblock \emph{arXiv preprint arXiv:2602.12675}, 2026.

\bibitem[Zhao et~al.(2026)Zhao, Zhu, Zheng, Zhou, Yan, Li, Yang, Li, and Zhu]{zhao2026causal}
Min Zhao, Hongzhou Zhu, Kaiwen Zheng, Zihan Zhou, Bokai Yan, Xinyuan Li, Xiao Yang, Chongxuan Li, and Jun Zhu.
\newblock Causal forcing++: Scalable few-step autoregressive diffusion distillation for real-time interactive video generation.
\newblock \emph{arXiv preprint arXiv:2605.15141}, 2026.

\bibitem[Zhao et~al.(2023)Zhao, Gu, Varma, Luo, Huang, Xu, Wright, Shojanazeri, Ott, Shleifer, et~al.]{zhao2023pytorch}
Yanli Zhao, Andrew Gu, Rohan Varma, Liang Luo, Chien-Chin Huang, Min Xu, Less Wright, Hamid Shojanazeri, Myle Ott, Sam Shleifer, et~al.
\newblock Pytorch fsdp: experiences on scaling fully sharded data parallel.
\newblock \emph{arXiv preprint arXiv:2304.11277}, 2023.

\bibitem[Zheng et~al.(2023{\natexlab{a}})Zheng, Lu, Chen, and Zhu]{zheng2023dpm}
Kaiwen Zheng, Cheng Lu, Jianfei Chen, and Jun Zhu.
\newblock Dpm-solver-v3: Improved diffusion ode solver with empirical model statistics.
\newblock \emph{Advances in Neural Information Processing Systems}, 36:\penalty0 55502--55542, 2023{\natexlab{a}}.

\bibitem[Zheng et~al.(2023{\natexlab{b}})Zheng, Lu, Chen, and Zhu]{zheng2023improved}
Kaiwen Zheng, Cheng Lu, Jianfei Chen, and Jun Zhu.
\newblock Improved techniques for maximum likelihood estimation for diffusion odes.
\newblock In \emph{International Conference on Machine Learning}, pages 42363--42389. PMLR, 2023{\natexlab{b}}.

\bibitem[Zheng et~al.(2025{\natexlab{a}})Zheng, Chen, Ye, Wang, Zhang, Jiang, Su, Ermon, Zhu, and Liu]{zheng2025diffusionnft}
Kaiwen Zheng, Huayu Chen, Haotian Ye, Haoxiang Wang, Qinsheng Zhang, Kai Jiang, Hang Su, Stefano Ermon, Jun Zhu, and Ming-Yu Liu.
\newblock Diffusionnft: Online diffusion reinforcement with forward process.
\newblock \emph{arXiv preprint arXiv:2509.16117}, 2025{\natexlab{a}}.

\bibitem[Zheng et~al.(2025{\natexlab{b}})Zheng, Chen, Chen, He, Liu, Zhu, and Zhang]{zheng2025direct}
Kaiwen Zheng, Yongxin Chen, Huayu Chen, Guande He, Ming-Yu Liu, Jun Zhu, and Qinsheng Zhang.
\newblock Direct discriminative optimization: Your likelihood-based visual generative model is secretly a gan discriminator.
\newblock \emph{arXiv preprint arXiv:2503.01103}, 2025{\natexlab{b}}.

\bibitem[Zheng et~al.(2025{\natexlab{c}})Zheng, Chen, Mao, Liu, Zhu, and Zhang]{zheng2025masked}
Kaiwen Zheng, Yongxin Chen, Hanzi Mao, Ming-Yu Liu, Jun Zhu, and Qinsheng Zhang.
\newblock Masked diffusion models are secretly time-agnostic masked models and exploit inaccurate categorical sampling.
\newblock In \emph{International Conference on Learning Representations}, volume 2025, pages 63186--63227, 2025{\natexlab{c}}.

\bibitem[Zheng et~al.(2025{\natexlab{d}})Zheng, Wang, Ma, Chen, Zhang, Balaji, Chen, Liu, Zhu, and Zhang]{zheng2025large}
Kaiwen Zheng, Yuji Wang, Qianli Ma, Huayu Chen, Jintao Zhang, Yogesh Balaji, Jianfei Chen, Ming-Yu Liu, Jun Zhu, and Qinsheng Zhang.
\newblock Large scale diffusion distillation via score-regularized continuous-time consistency.
\newblock \emph{arXiv preprint arXiv:2510.08431}, 2025{\natexlab{d}}.

\bibitem[Zhou et~al.(2024)Zhou, Zheng, Wang, Yin, and Huang]{zhou2024score}
Mingyuan Zhou, Huangjie Zheng, Zhendong Wang, Mingzhang Yin, and Hai Huang.
\newblock Score identity distillation: Exponentially fast distillation of pretrained diffusion models for one-step generation.
\newblock In \emph{Forty-first International Conference on Machine Learning}, 2024.

\bibitem[Zhu et~al.(2026)Zhu, Zhao, He, Su, Li, and Zhu]{zhu2026causal}
Hongzhou Zhu, Min Zhao, Guande He, Hang Su, Chongxuan Li, and Jun Zhu.
\newblock Causal forcing: Autoregressive diffusion distillation done right for high-quality real-time interactive video generation.
\newblock \emph{arXiv preprint arXiv:2602.02214}, 2026.

\bibitem[Zou et~al.(2026)Zou, Zheng, Liu, Hang, Liu, and Yu]{zou2026hiar}
Kai Zou, Dian Zheng, Hongbo Liu, Tiankai Hang, Bin Liu, and Nenghai Yu.
\newblock Hiar: Efficient autoregressive long video generation via hierarchical denoising.
\newblock \emph{arXiv preprint arXiv:2603.08703}, 2026.

\end{thebibliography}
